\documentclass[a4paper,11pt]{report}

\usepackage[latin1]{inputenc}
\usepackage[english]{babel}
\usepackage{graphicx}
\usepackage{amsmath}
\usepackage{amsfonts} 
\usepackage{fontenc}
\newcommand{\HRule}{\rule{\linewidth}{0.2mm}}
\begin{document}
\begin{titlepage}

\begin{center}
\includegraphics[width=0.35\textwidth]{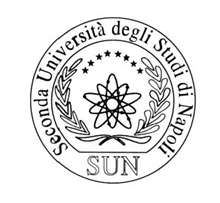}\\[1cm]    
\textsc{\LARGE Seconda Università degli Studi di Napoli}\\[1.5cm]
\textsc{\Large Facoltà di Ingegneria}\\
\textsc{Corso di Laurea Magistrale in Ingegneria Informatica}
\\
\textsc{Dipartimento di Ingegneria dell'Informazione}\\
[0.5cm]

\HRule \\[0.4cm]
{ \huge \bfseries A behavioural approach to obstacle avoidance for mobile manipulators based on distributed sensing}\\[0.4cm]

\HRule \\[1.5cm]

\begin{minipage}{0.4\textwidth}
\begin{flushleft}\large
\emph{Relatore:}\\
{Ch.mo Prof.~Ciro Natale}
\end{flushleft}
\end{minipage}
\begin{minipage}{0.4\textwidth}
\begin{flushright}\large
\emph{Candidato:} \\
Luigi Palmieri\\
A18\slash 016
\end{flushright}
\end{minipage}

\vfill


{Anno Accademico 2010\slash 2011}
\end{center}

\end{titlepage}

\newpage

\begin{flushright}
\textit{ alla mia famiglia}

\textit{ ai miei amici }
\end{flushright}

\newpage

\begin{flushright}
    \textit{Ma chi opera la verità viene alla luce, perchè appaia chiaramente che le sue opere sono state fatte in Dio. Gv 3,21 }
\end{flushright}

\begin{flushright}
    \textit{Non basta guardare, occorre guardare con occhi che vogliono vedere, che credono in quello che vedono. Galileo Galilei }
\end{flushright}

\newpage
\begin{flushright}
	\textit{Ringraziamenti}
\end{flushright}

Credo che non sia semplice individuare delle persone da ringraziare per la stesura di questa tesi, in quanto essa è frutto di confronti e studi attinenti alle tematiche affrontate, ma anche frutto di motivazioni e suggerimenti derivati dai non tecnici. A valle di ogni mio impegno posso dire che la tesi è sintesi di quanto coltivato lungo tutto il percorso universitario: conoscenze, incontri, scontri e soprattutto avere cura di una passione che nasce attraverso essi.

Inizio con il ringraziare il Prof. Ciro Natale, per la sua costante disponibilità e dedizione al confronto, oltre che ai continui consigli sul come affrontare alcune difficoltà emerse. Egli è stato, senza alcun dubbio, un'ottima guida per la realizzazione di questo lavoro di tesi, sia per il suo impegno che per le sue conoscenze.
Ringrazio i professori del laboratorio di Automatica De Maria, Cavallo e Pirozzi e il dottore di ricerca Falco, perchè è anche grazie al loro impegno ed esempio nella didattica e nel laboratorio, che ho coltivato la passione per la robotica e per l'automatica in genere. Ovviamente i ringraziamenti si estendono anche ai restanti professori conosciuti lungo il percorso universitario: ogni esame è stato costruttivo e fondamentale per la mia formazione.

Arriviamo dunque ai ringraziamenti non tecnici. I tanti esami sono stati affrontati con il continuo studio, quasi mai solitario, ringrazio dunque tutti i miei amici dell'università con cui ho condiviso sogni, esami, progetti e risate varie.  

Ringrazio gli amici, quelli che ti ascoltano, che fanno di tutto per non perderti, di quelli che non dicono paroloni e non fanno enormi promesse, ma sai che puoi contare su di loro.

Ringraziare i genitori è necessario ma è anche una meraviglia: rendersi conto che riesci ad essere costante nell'impegno e ad avere creatività perchè c'è qualcuno che ti ha insegnato ad essere così è fondamentale, anche se l'insegnamento non è mai arrivato per via di metodi didattici. A mio padre e mia madre poi si affiancano i miei fratelli, insostituibili presenze.

Un ultimo ringraziamento va a Lui, non si \emph{gioca a dadi} qui, c'è il Suo sogno per ognuno di noi, a noi non resta che viverlo.

\begin{flushright}
    \emph{A voi tutti il mio Grazie.}
\end{flushright}

\newpage
\tableofcontents

\chapter{Introduction}\label{c:primo}
There was a time when the robots were thought in the Asimov's mind as intelligent  mechanical structures that obey to the following three laws: 
\begin{enumerate}
    \item \textit{A robot may not injure a human being or, through inaction, allow a human being to come to harm};
 \item \textit{A robot must obey orders given it by human beings except where such orders would conflict with the First Law};
\item \textit{A robot must protect its own existence as long as such protection does not conflict with the First or Second Law}.
\end{enumerate}
 
In Asimov's science fiction vision there was already the presence of two important aspects of the robotics: to give orders and safety.  

According to the trends of the robotics research in a few of decades, robots will be able to share the same dynamically changing environments with the human beings and these latter will not be worried to share the same place with some \emph{mechanical friends}. To arrive to this scenario the research has to move its point of view. Industrial robots, well-known as manipulators, have been developed thinking them as operators in a cell, where they do some planned tasks, moving and rotating the end-effector following a predetermined path. 

Industrial robots work in a separate cell where no human beings can be, according to safety standards. \emph{Advanced robotics} moves away from this view. The robot has to share the same human environment. To afford this idea, the research has to re-think the way how a robot interacts with the environment and a human.  A robot has to become an autonomous system: it has to accomplish some tasks interacting independently of the surroundings . Furthermore, a robot needs to cooperate with human beings without damaging their lives; the research needs a new approach to the interaction between humans and robots, finding a way to develop machines that act like humans (in behaviours and in physical structures) and use low forces, similar to the human ones. 

In this way, a robot can be used not only on an assembly line but in many places, also the ones where humans cannot work, for example in a critical situation of a nuclear central where it is dangerous to expose people to the radioactivity, while for the robot could be less dangerous, this is just one example; we could think also to space robots or underwater robots. There are already some robots used in the cited above contexts.

Service robotics is also a very useful application of robotics: the robots are seen as  helpers in the human activities, for example in the domestic environments for cleaning operations, or to help erderly people. 

In  \emph {advanced robotics}, one of the most important issues to address is to equip the robots with the capabilities to react to human actions or even take the initiative to interact with them, by taking decision on the basis of data coming from the sensors and background knowledge.

 The most important areas of interest of the advanced robotics research are: actuation and power systems, haptic and interaction technologies, artificial intelligence and learning, sensors fusion, motion planning. A new way to design actuation and power systems has to be found especially to improve energy savings and to get movements similar to those of humans: why a robot cannot move like a human?
 Haptic and interaction research tries to develop the sense of touch of the haptic interfaces and to improve the way how the robot reacts to the sense of touch.

 As described before, in advanced robotics the sensor system is the base to tackle some problems. According to the kind of the sensors that you have got on your robot, you can perform different tasks. The information given by the sensors are not always of the same type, so they need to be filtered in a way to extrapolate the right information: this is the objective of the sensor fusion. 

Within motion planning, \emph{an important task to obtain is to avoid an obstacle} situated nearby the path that the robot has to follow. How can we achieve this objective? A human being would move away to protect all the parts of his body, moving it without hitting other obstacles. Then, when the human being passes the obstacle, he goes back to the action he was doing before the obstacle came. Until today the research tried to pursue the same behaviours on mobile robots.

 A mobile robot, equipped with proximity sensors, would act similarly but not perfectly like a human being does. The main limitations currently are: until today a robot does not find a way to prevent itself from the obstacle in a global way and furthermore it deviates from the obstacle without a global perception of the surroundings.

 As it will be showed in this work (\emph{2nd Chapter}) the robotics research has elaborated some techniques to achieve the obstacle avoidance but all of them do not think of a robot in its completeness, they compute commands to move away from the obstacle as it was a single piece: few sensors calculate the distances between the robot and the obstacles and according to these measures an algorithm elaborates the commands.  

Few sensors can permit to obtain only a robot conservative representations in the environment, they allow to approximate the position of a robot as a geometric primitive in the space. Even more so, for a manipulator what the research has done is to think of it in a conservative way, due to the larger complexity related to the high number of degrees of freedom describing its configuration.

This work tries to elaborate a new approach to the obstacle avoidance: why a robot has to accomplish in a conservative way the obstacle avoidance? Is it possible to think to obstacle avoidance based on  \emph{distributed} proximity sensing? 

 In the \emph{3rd Chapter} we try to describe the reasons to the need of a new approach to obstacle avoidance: a behavioural approach based on distributed proximity sensing. The \emph{4th Chapter} describes how we implemented the new approach, we used two toolboxes of Simulink: SimMechanics and the VR toolbox.

Afterwards in the \emph{5th Chapter} conclusions are drawn describing the results and some possible future developments.
\newpage
\chapter{State of the Art} \label{c:secondo}

Real-time obstacle avoidance is a prerequisite for Human-Robot Interaction, for service robotics and for all the robotic applications working into a dynamically changing environment specifically for autonomous robotic systems. Generically, the obstacle avoidance is a task to achieve together with other tasks, we could think to a manipulation task: while the robot is doing its job we would like to allow the avoidance of an incoming unpredicted obstacle. Furthermore obstacle avoidance actions can include the self-collision avoidance concept namely to permit to a manipulator, a mobile robot or a humanoid robot to avoid the collision between its links. Past research has already found ways to get the obstacle avoidance. 
In dynamically changing environments what really matters is a way to get advantage of the kinematic redundancy and the ability to merge single elementary behaviours to compose a complex one: Null-Space-Based behavioural control does this using a projection mechanism. We must give the same importance to the algorithms to obtain the obstacle avoidance: they  have been divided into motion re-planning algorithms and reactive control algorithms. 

Below is reported a brief review of the state of the art of planning algorithms, reactive control algorithms and behavioural-based control.

\section{Planning Algorithms}
\begin{figure}[ht]
\centering%
\includegraphics[height=150pt]{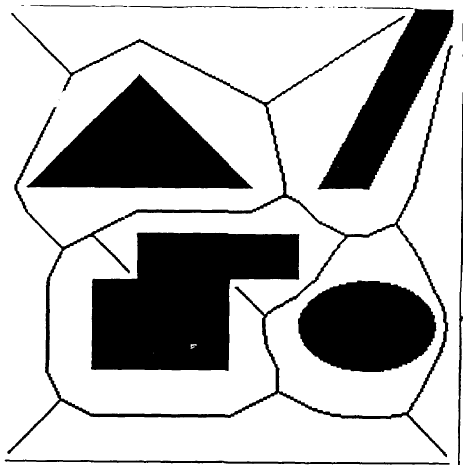}%
\caption{A generalized Voronoi Diagram. \label{fig:diag_voronoi}}%
\end{figure}

 The first kinds of re-planning algorithms used were the Road maps, where the planning algorithms use a retraction method based on the generalized Voronoi Diagrams \cite{rif1} (Figure \ref{fig:diag_voronoi}). Consequently a cell decomposition algorithm, which finds a path on a connectivity graph, was adopted \cite{rif2}. Re-planning algorithms were interpreted also in a probabilistic way, trying to explore the free configuration space, near the robot, using a space sampling. 

\begin{figure}[ht]
\centering%
\includegraphics[height=180pt]{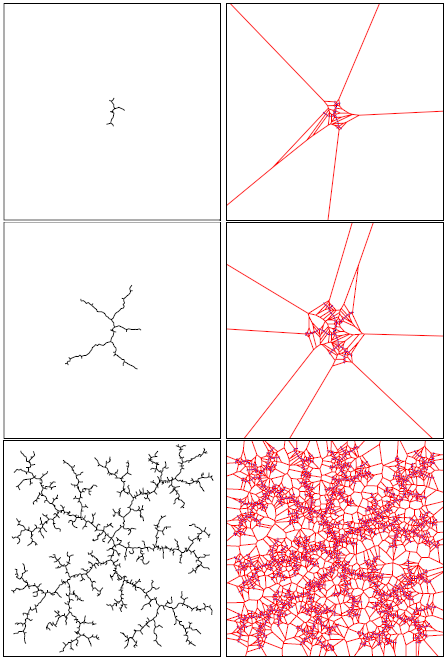}%
\caption{An RRT is biased by large Voronoi regions to rapidly explore, before uniformly covering the space. \label{fig:RRT}}%
\end{figure}

There are two types of algorithms: the Probabilistic Road Map (\emph{PRM}) a multiple-query stochastic method \cite{rif3}, and the Rapidly-exploring Random Tree (\emph{RRT}) \cite{rif4} (Figure \ref{fig:RRT}) a single-query stochastic method, where the free configuration space is explored only in the interesting parts.

 All the previous mentioned methods are planning algorithms very useful in static environments, they are executed in the planning layer of the robot. These are performed obviously thanks to the information received by the sensors. 

\section{Reactive Control Algorithms: The Artificial Potential}
\begin{figure}[ht]
\centering%
\includegraphics[height=180pt]{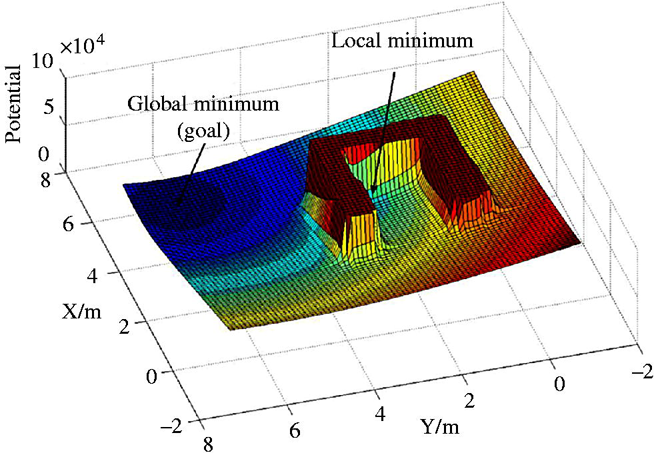}%
\caption{An example of a potential field with local and global minimum. \label{fig:pfield}}%
\end{figure}

A different approach is to avoid the obstacles directly using a control algorithm. We found in the literature different methods. They are based on a reactive control resulting from the computation of an artificial potential (an example in Figure \ref{fig:pfield}) . A very important improvement was done by the introduction of the repulsive forces, due to an artificial potential, acting as forces in the operational space of a robot. Citing Khatib in his work \cite{rif5} : 

\begin{quote}
\emph{The manipulator moves in a field of forces. The position to be reached is an attractive pole for the end effector and obstacles are repulsive surfaces for the manipulator parts}.
\end{quote}

The control of manipulators in  operational space is based on the choice of the forces F as control inputs. $\tau$ are the corresponding torques (or forces) applied to the joint actuators. The relationship between forces and joint torques (or forces) is the following:

\begin{equation}\label{opspace}
    \uppercase{\tau}=J^{T}(q)F
\end{equation}

The key idea of the method is to compute the repulsive forces due to the artificial potential, which consists generically of two terms:

\begin{equation}
    U_a(x)=U_d(x) + U_o(x)
\end{equation}

$U_d(x)$ is the artificial potential that attracts the robot to the desired configuration. $U_o(x)$ is the non-negative continuous and differentiable function that creates at each point a potential to reject the manipulator when it is too near to the obstacle. The forces applied are:

\begin{equation}
    F_a(x)=F_d(x) + F_o(x),
\end{equation}
 where
\begin{equation}
    F_d(x)= -\nabla[U_d(x)],\space 
    F_o(x)= -\nabla[U_o(x)].
\end{equation}
To obtain asymptotic stabilization of the system he added  dissipative forces proportional to $\dot{x}$.
\begin{equation}
    F_m(x)= -k_p(x-x_d)-k_v\dot{x}
\end{equation}

Khatib in his work used the following potentials:

\begin{equation}
    U_d(x)=\frac{1}{2}k_p(x-x_d)^{2} \\
\end{equation}

\begin{equation}\label{rep}
U_o(p) =\left\{
\begin{split}
&\frac{1}{2}\eta\left(\frac{1}{\rho}-\frac{1}{\rho_o}\right)^{2}, \; \hbox{if} \; \rho \le \rho_o \\ &0, \hbox{otherwise}
\end{split}
\right.
\end{equation}

where $\rho_o$ represents the limit distance of the potential field and $\rho$ is the shortest distance to the obstacle. In this approach it is clear that from all of the points subjected to the potential, it is selected the one with the \emph{shortest distance to the obstacle}.
The control is achieved using the following force (it is equal to zero when $\rho>\rho_o$ ) considering a single obstacle:

\begin{equation}
    F_o(x)=\eta\Bigg(\frac{1}{\rho}-\frac{1}{\rho_o}\Bigg)\frac{1}{\rho^{2}}\frac{\partial\rho}{\partial \textbf{x}}
\end{equation}
where

\begin{equation}
    \frac{\partial\rho}{\partial \textbf{x}}=\Bigg[\frac{\partial\rho}{\partial x } \frac{\partial\rho}{\partial y} \frac{\partial\rho}{\partial z} \Bigg]^{T}
\end{equation}

In his approach Khatib outlined that the obstacle avoidance problem  could be divided in two stages:
\begin{itemize}
    \item at high level control, generating a global strategy for the manipulator's path in terms of intermediate goals;
	\item at the low level, producing the appropriate commands to attain each of these goals, taking into account the detailed geometry and motion of manipulator and obstacle, and making use of real-time obstacle sensing.
\end{itemize}

The complexity of tasks that can be implemented with this approach is limited. In a cluttered environment, local minima can occur in the resultant potential field. This approach has a local perspective of the robot environment, the robot may follow in a stable position before reaching its goal.

According to Khatib's vision any point of the robot can be subjected to the artificial potential field. The superposition property (additivity) of potential fields enables the control of a given point of the manipulator with respect to this obstacle by using the sum of the relevant gradients. Per each obstacle $O_i$ described by a set of primitives $P_p$ we have:
\begin{equation}
F_{O_i,psp}^{*}=\sum_pF_{(P_p,psp)}^{*}
\end{equation}

Alternatively we can have a reactive control algorithm in a kinematic way, just mapping the gradient of the total potential in the joint velocity of the manipulator using the transposed Jacobian.

Let $q$ be the configuration of the robot, $p_i(q)$ (for $i=1,.. P-1$) is a control point situated on the manipulator where the potential are calculated. The control point for the end effector is $P_p$. The velocities applied to the manipulator are the following:

\begin{equation}
\dot q = - \sum_{i=1}^{P-1} J_i^T(q) \nabla U_r(p_i) -J_P^T(q)\nabla U_t(P_p)
\end{equation}

While Khatib proposed the repulsive potential (the one in equation \eqref{rep}) other types of potential were developed.

\subsection{The Harmonic Potential}
 The main idea behind the Harmonic Potential method (\cite{rif6}) is to have an artificial potential, without local minima, using the Laplace equation. The solutions of the Laplace equation are called \emph{harmonic} functions. 

In the real world, many physical problems are described by the Laplace equation. An example is the incompressible fluid, a steady state electric charge distribution and  also a steady state temperature distribution also follows the Laplace equation.
 The equations regulating the physical behaviours of an incompressible fluid are:

\begin{equation}
\nabla^{2} \phi(x) =0,  x \epsilon R^{3}
\end{equation}
From which we obtain the following system:
\begin{equation}
\dot{x}=-\nabla\phi(x),x(0)=x_o \epsilon R^{3}
\end{equation}

The first important property of a harmonic function is the principle of superposition, which follows from the linearity of the Laplace equation. That is,$\phi_1$ and $\phi_2$ are harmonic, then any linear combination of them is also harmonic and a solution of the Laplace equation.

A drawback of HPFs is that they cannot be directly used for generating virtual driving forces as in Khatib work is done. 

In this case a potential gradient is replaced by velocity field.

\subsection{Circular Fields}
\begin{figure}[ht]
\centering%
\includegraphics[height=180pt]{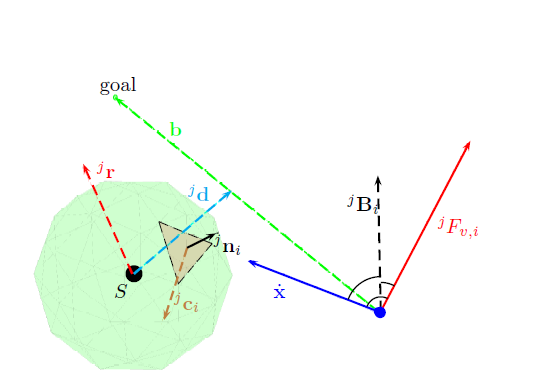}%
\caption{Principle of the current definition for Circular Fields for obstacle j with surfaces i and robot n. \label{fig:circ}}%
\end{figure}

 Another important approach to artificial potentials is the Circular Fields algorithm.
The physical phenomena behind the Circular Fields (Figure \ref{fig:circ}) is the generation of \emph{N} artificial electromagnetic-fields by the currents circulating in \emph{N}  wires, of indefinite length, situated on the robot surfaces \cite{rif7}.
The forces applied to the robot are the following:
\begin{equation}
F_j = \dot{x} \times \sum_{i=1}^N B_i
\end{equation}
\begin{equation}
B_i= I_k \frac{c_j \times \frac{\dot{x}}{\left\|x\right\|}}{l_i^2}d_{i}
\end{equation}
where $I_k$ is the virtual current factor, $c_j$ is the current direction vector of surface element i, $d_i$ is the surface element area, $l_i=\left\|x-x_i\right\|$ is the distance of the current position of the point mass x, and $x_i$ is the position of the obstacle surface element i of obstacle j.
Ultimately, a few modifications have been proposed for this approach \cite{rif8}.

The algorithms presented above are all used as part of the robot control algorithm. The next presented algorithm merges planning with the forces obtained by an artificial potential.

\section{Elastic Strips}

\begin{figure}[ht]
\centering%
\includegraphics[height=180pt]{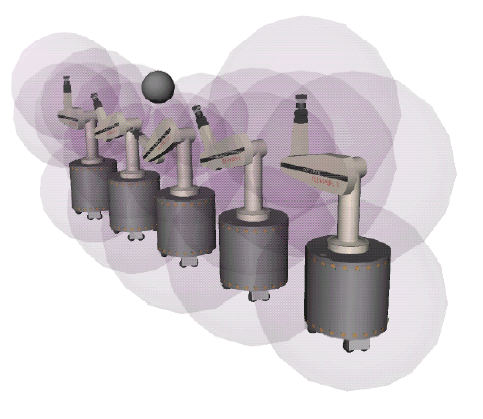}%
\caption{Elastic tunnel in the presence of an obstacle. \label{fig:tunnel}}%
\end{figure}

The Elastic Strip \cite{rif9} is a framework used to merge reactive motion to the classical motion planning. The framework defines an \emph{elastic tunnel} that contains the \emph{candidate path} generated by a planner. Virtual Robots move on homopatic path computed from the candidate path using a potential field-based control algorithm. The virtual robot moves in the elastic tunnel (example in Figure \ref{fig:tunnel}) as it was guided by strings: one to repulse the robot from an obstacle situated under a distance threshold, and others to permit the robot to go back to the candidate path if the obstacle would recede. Differently from other types of motion planning algorithm, in this framework there is not the need to explore the entire configuration space, instead it maps proximity information from the environment into the configuration space, using the kinematics of the manipulator.

\begin{figure}[ht]
\centering%
\includegraphics[height=150pt]{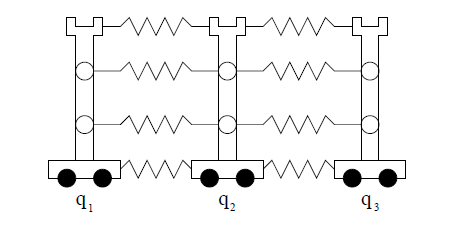}%
\caption{Principal structure of elastic strip. \label{fig:strips}}%
\end{figure}%

To define the elastic strip we have to add incrementally to the candidate path the path generated by the forces calculated from an internal and an external potential.
The external potential used is the following:

\begin{equation}
V_e(p) =\left\{
\begin{split}
&\frac{1}{2}k_r(d_o-d(p))^{2}, \; \hbox{if} \; d(p)<d_o \\ &0, \hbox{otherwise}
\end{split}
\right.
\end{equation}
where $d(p)$ is the distance from $p$ to the closest obstacle, $d_o$ defines the region of influence around obstacles, and $k_r$ is the repulsion gain. The force that acts on a single point p (the closest one) is the following:

\begin{equation}
F_p^{ext}=-\nabla V_{e} = -k_r(d_o-\left\|d(p)\right\|)\frac{d}{\left\|d(p)\right\|},
\end{equation}
where \emph{d} is the vector between \emph{p} and the closest point on an obstacle. 
The internal force, corresponding to the inertial potential is:
\begin{equation}
F_{i,j}^{int}=kc\Bigg(\frac{d_j^{i-1}}{d_j^{i-1}+d_j^i}(p_j^{i+1}-p_j^{i-1})-(p_j^i-p_j^{i-1})\Bigg)
\end{equation}
where $d_j^i$ is the distance $\left\|p_j^i-p_j^{i+1}\right\|$ in the initial path and $k_c$ is the contraction gain of the elastic strip (its structure is depicted in Fig. \ref{fig:strips}).
The robot moves through the new path contained in the elastic strips. The forces are never applied directly to the robot but they are mapped to joint displacements using the kinematic model of the manipulator. This effectively replaces configuration space exploration with a directed search, guided by work space forces.

In case of redundant manipulators, a third potential (\emph{$V_p$}) is introduced. The potential $V_p$ is associated to the posture, it can be used to define a preferred posture for the robot in absence of other constraints.

 The integration of planning and control allows to delay the exploration of the configuration space necessary for plan generation until execution, executing re-planning more efficiently.
The elastic strips is applied to a single point (for a manipulator, for each link there is a single point, chosen as the one with the shortest distance to the obstacle). This approach is a good strategy to solve the problem of the local minima because it is a local and global method in the same time, but elastic strips cannot replace planning itself: if changes in the environment are substantial and the elastic strip framework cannot find a valid candidate path, the re-invocation of a complete planner becomes necessary.

\section{The skeleton algorithm}

The skeleton algorithm in \cite{rif10} is an important step forward in reactive collision avoidance because it outlines a way to compute the position of the control points and the corresponding Jacobian matrices needed for control. The algorithm, developed for the self-collision avoidance, is composed of these four steps:
\begin{itemize}
 \item building a proper model of the robot, namely the skeleton, useful for analytical computation;
\item finding the \emph{closest point} to a possible collision along the skeleton, namely the collision points;
\item generating repulsion forces;
\item computing avoidance torque commands to be summed to the nominal torques for the controller.
\end{itemize}

Once built, the model of the robot, which in general can be built by hand or it could be derived (\emph{the skeleton}) automatically from a proper kinematic description (e.g via a Denavit-Hartenberg table), we have to find the collision points. They move along the segments of the skeleton. Hence, the direct kinematics computation can be carried out in a parametric way for a generic point on each segment by simply replacing the link length in the homogeneous transformation, relating two subsequent frames with the distance of the collision point from the origin of the previous frame. 

The collision points $p_{a,c}$ and $p_{b,c}$ are found computing the minimum distance between the two segments. To the computation of the avoidance torques, it is necessary to compute the Jacobians associated with the collision points, i.e. the matrices $J_{a,c}$ and $J_{b,c}$ describing the differential mapping of $\dot{p}_{a,c}$ and  $\dot{p}_{b,c}$, with the joint velocities $\dot{q}$ of the whole structure, i.e. in compact form:
$$
\left[
\begin{array}{c}
\dot{p}_{a,c}\\
\dot{p}_{b,c}\\
\end{array}
\right] =
\left[
\begin{array}{c}
J_{a,c} \\
J_{b,c} \\
\end{array}
\right]
\dot{q}
$$

We have to notice that the positions of $p_{a,c}$ and $p_{b,c}$ vary along the segments as the manipulator is moving.

This algorithm calculates the forces as repulsion forces of an artificial potential
\begin{equation}
U_c(p)=\left\{
\begin{split}
&\frac{1}{2}k(d_{start}-d_{min}+d_o)^{2}, \; \hbox{if} \; d_{min}<d_o+d_{start} \\ &0, \hbox{otherwise}
\end{split}
\right.
\end{equation}
where \emph{$d_{min}$} is the minimum distance between two collision points, \emph{$d_{start}$} is the starting distance where the force has to act: points farther than \emph{$d_{start}$} are not subject to any repulsion. Moreover, \emph{$d_o$} is the limit distance around the skeleton where a collision may occur: in the case of cylindrical links, $d_o$ is the radius of the section of the link. The following forces are defined if, $d_{min}<d_o+d_{start}$ :

\begin{equation}
f_{a,c}=\frac{k(d_{start}-d_{min}+d_o)}{d_{min}}(p_{a,c}-p_{b,c}) - D_a\dot{p}_{a,c}\end{equation}
\begin{equation}
f_{b,c}=\frac{k(d_{start}-d_{min}+d_o)}{d_{min}}(p_{b,c}-p_{a,c}) - D_b\dot{p}_{b,c}
\end{equation}
where $D_a$ and $D_b$ are positive definite matrices. Multiple collision points on the same segment may need to be considered, whenever more segments are close to a possible collision. The damping matrix is used to increase the values of forces applied to the collision points for the presences of multiple collision points.

The last step of the algorithm is to compute the \emph{avoidance torques} corresponding to the repulsion forces via the Jacobian transpose defined above i.e : 
$$
\tau_c =\left[
\begin{array}{c}
J_{a,c}^{T} J_{b,c}^{T}
\end{array}
\right] \left[
\begin{array}{c}
f_{a,c} \\
f_{b,c} \\
\end{array}
\right]
$$

These torques are added to the ones used for the control. As an alternative to the presented technique, the skeleton algorithm could be applied also for a velocity control based implementation.

\section{The Null-Space-based behavioural control}
Behaviour-based approaches are methodologies to design the control architecture of artificial intelligence systems. The key idea of behaviour-robotics is that intelligence of the robotic system is provided by a set of behaviours, designed to achieve specific goals, that are activated on the basis of sensor information. A behaviour is expressed through a function of the robot configuration that measures the degree of fulfilment of the task.

There are different approaches to handle multiple elementary tasks to be executed simultaneously. The solutions to this problem of \emph{behavioural coordination} can be divided in two categories: \emph{competitive methods} and \emph{cooperative methods}.

In the competitive methods the coordination can be viewed as a competition among behaviours; only one behaviour wins and only its response  is sent to the robot for execution. An example of competitive methods is the layered architecture proposed in \cite{rif11}. Layers have different priority levels and a hierarchy solves the conflict among the tasks, higher-level tasks can subsume the lower-level ones: it is also known as \emph{subsumption architecture}. 
Cooperative methods merges the output of more than one behaviour at a time. A supervisor gives as output a solution, calculated as the sum of all motion commands (one for each task) opportunely multiplied by a gain vector, which can dynamically change on the basis of sensor information: the \emph{motor schema control} (\cite{rif12}) is an example of cooperative behaviour coordination. 
Differently from the competitive approach in the cooperative one the output is a linear combination of all behaviours, all of them participate to the control of the robot. 

An important step forward is done by the \emph{Null-Space-based Behavioural} (NSB) control (\cite{rif13}). This technique is based on some properties of the inverse kinematics (an overview is reported in \cite{rif14}). For a redundant manipulator the inverse kinematics problem admits an infinite number of solutions and a criterion to select one of them is needed.

 A lot of work has been done considering the problem of redundancy resolution at the inverse differential kinematics level. The simplest method is based on the use of the pseudo-inverse of the Jacobian matrix: this guarantees optimal reconstruction of the desired end-effector velocity -in a least-squares sense- with the minimum-norm joint velocity. Redundancy can be exploited to meet constraints additional to the solution of the inverse kinematics problem. A possible constraint is the obstacle avoidance. Generically, a term is added for local optimization of a scalar criterion, i.e a term proportional to the gradient of the criterion  is projected onto the null space of the Jacobian matrix so that the  end-effector task is not affected.

The task-space augmentation is another approach to the resolution of the redundancy. It introduces a constraint task to be fulfilled along with the end-effector task, so an augmented Jacobian matrix is set-up whose inverse gives the sought joint velocity solution.

 A stone on which the NSB is based is the framework of the task-priority strategy (\cite{rif15}, \cite{rif16}). In this approach  the conflicts between the end-effector task and the constraint task is resolved by assigning an order of priority to the given tasks and then satisfying the lower priority task only in the null space of the higher priority task. 

In the NSB for each behaviour we need a task variable $\sigma \in R^{m}$, and we define $p \in R^{n}$ the system configuration, then we have:
\begin{equation}
\sigma = f(p) 
\end{equation}
with the corresponding differential relationship:
\begin{equation}
\dot{\sigma} = \frac{\partial{f(p)}}{\partial{p}}\dot{p}=J(p)v
\end{equation}
where $m$ is the task function dimension, $J \in R^{m \times n}$ is the configuration dependent task Jacobian matrix, and $v := \dot{p} \in R^{n \times 1}$ is the system velocity. Notice that the only case of interest is $m \le n$; otherwise if $m > n$, the task would be unfeasible or the null space of a full rank $J(p)$ would be empty thus preventing the possibility of controlling any other task.

Consider a generic behaviour $k$ defined by the task variable $\sigma_k$ having a desired value $\sigma_{d,k}$ and a Jacobian $J_k$. The reference velocity of the generic $k^{th}$ task can be calculated by the Closed Loop Inverse Kinematics Algorithm (CLIK) as
\begin{equation}
v_k = J_{k}^{\dagger } ( \dot{\sigma}_{d,k} + \Lambda_k \tilde{\sigma_k} ),
\end{equation}
where $J_{k}^{\dagger } = J_k^{T}(J_k J_k^{T})^{-1}$ (when $m \le n$ and  $rank(J_k(p))=m$), $\Lambda_k$ is a suitable constant positive-definite matrix of gains and $\tilde \sigma_k$ is the task error defined as $\tilde \sigma_k=\sigma_{d,k} - \sigma_k$. The term $ \Lambda_k \tilde{\sigma_k} $ is added to counteract the numerical drift due to discrete-time integration (\cite{rif17} describes how to choose $\Lambda_k$ to ensure the stability of the algorithm).

A velocity vector for each behaviour is computed as if it was acting alone; then, before adding the single contribution to the overall vehicle velocity, a lower-priority behaviour is projected onto the null space of the higher-priority behaviours so as to remove those velocity components that would conflict with it. An example of the merging of three tasks could be:
\begin{equation}
\label{eq_nsb}
v_d = v_1 + (I - J_1^{\dagger}J_1)[v_2 + (I-J_2^{\dagger}J_2)v_3]
\end{equation}

Equation (\ref{eq_nsb}) has a geometrical interpretation. Each task velocity is computed as if it were acting alone; then, before adding its contribution to the overall robot velocity, a lower-priority task is projected onto the null space of the immediately higher-priority task so as to remove those velocity components that would conflict with it. 

A supervisor might be considered in order to dynamically change the relative task priorities. The research until today does not suggest the type of the supervisor to use and at the same time does not give any condition of stability for it.  The lower-priority tasks are fulfilled only in a subspace where they do not conflict with the ones having high priority. This is clearly an advantage with respect to the competitive approaches, where one single task can be achieved at once, and to the cooperative approaches, where the use of a linear combination of each single task's output has as result that no single task is exactly fulfilled and furthermore the designer needs to heuristically tune the parameters.

In the case of multiple non-conflicting tasks the NSB does not guarantee that the lower priority  tasks is instantaneously achieved with the sub-optimal velocity. Nevertheless, in the considered case, the closed loop ensures that the error of secondary tasks converges to zero.

\section{On distributed sensor systems}

A distributed sensor system is generally thought as an array of sensors. They are distributed on a surface where we want to collect some information. In order to obtain the obstacle avoidance we focus our attention on proximity sensors. We might think to a robot skin  that needs to be covered with a sensor array to detect proximity of an object. The proximity sense is the sense that tell us how far from the skin of the robot is an object. A proximity sensor complements the gap between vision sensors and tactile sensors: vision sensors can see wide areas but sometimes occlusion occur; tactile sensors are useful to detect a contact on a surface.
An impressive example of distributed proximity sensor system is the "Net-Structure Proximity Sensor" developed by  the UEC Shimojo Laboratory in Tokio \cite{rif18}.
In this sensor, infrared LED and photo-transistors are used as the detection elements. The sensor outputs the center position of the nearby object and the distance from the sensor surface.The sensor can be laid out like a net to cover arbitrary surfaces. Response time is less than 1 ms without regard to the number of detection elements.
The circuit diagram (see Figure \ref{fig:sensors}) of Net-Structure Proximity Sensor are shown below. An infrared LED is allocated in a set with each of the photo-transistors. The infrared light emitted from the LED reflects on a nearby object, and then is radiated to the array of photo-transistors. Here, current distribution occurs in the circuit according to the difference of the amount of the light received by the photo-transistors. 
By calculating the first moment of the current distribution from voltages on four terminals (Vsl, Vs2, Vs3, Vs4), the center position of the object is obtained.
Furthermore, since the amount of the light received on the photo-transistors changes according to the distance between the sensor surface and the object, the distance (z) is also estimated from the total amount of the current in the circuit.
\begin{figure}[ht]
\centering%
\includegraphics[height=180pt]{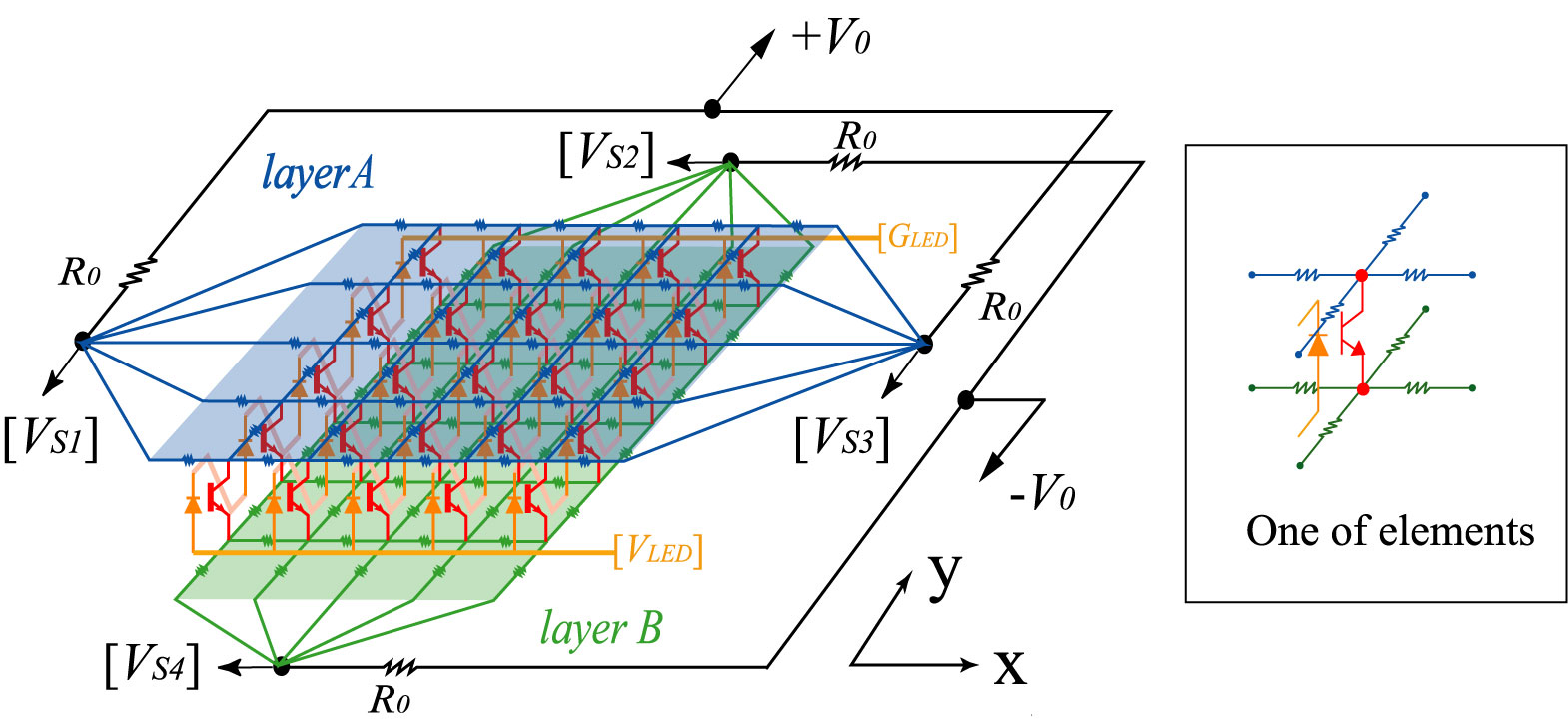}%
\caption{Net-Structure Proximity Sensor. \label{fig:sensors}}%
\end{figure}%

\begin{figure}[ht]
\centering%
\includegraphics[height=180pt]{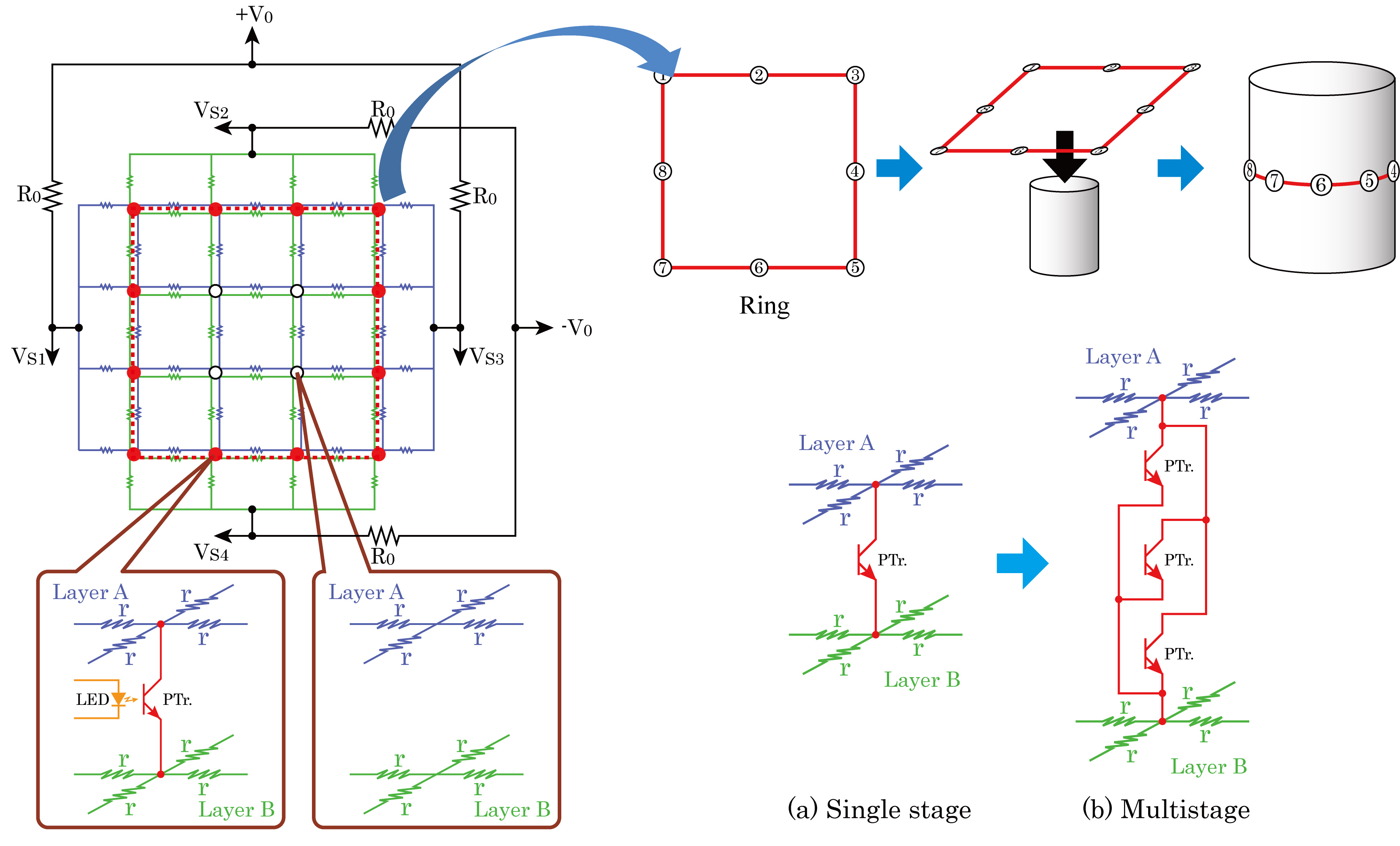}%
\caption{Ring-Shaped Proximity Sensor. \label{fig:sensori}}%
\end{figure}

Moreover the same group has developed still a new design of Net-Structure Proximity Sensor. The basic structure is the same as the Net-Structure Proximity Sensor, but photo-transistors are arranged on the edges of the m x n matrix as indicated in figure \ref{fig:sensori}.

This sensor is implemented on a cylindrical surface. Each photo-transistor faces outside in radial direction. In this geometric arrangement, the output of the center position (xc,yc) corresponds to the direction of a nearby object, and so the sensor can detect 360 degrees all around seamlessly. This sensor is called \emph{Ring-Shaped Proximity Sensor} .

\chapter{A new  distributed approach to the obstacle avoidance}\label{c:terzo}

\section{Research Question}

As anticipated in the introduction, the objective of this work is to develop an algorithm for the obstacle avoidance of a mobile manipulator in a dynamically changing environment. The aim is to let the robot autonomously  move away from the obstacles situated on its path. What the robotics research has done until today is to achieve this task in a conservative way, considering the robot as a single body and finding a way to escape from the obstacle. As the \emph{Chapter \ref{c:secondo}} showed, there are many methods to achieve this objective, but these methods act in a conservative way: 
\begin{itemize}
    \item the \emph{artificial potential} sees the robot subjected to a field of forces, computed in few so called control points situated along the manipulator, the force field's gradients are computed and then projected into the transposed Jacobian;
	\item the \emph{elastic strips} method permits the robot to move in a strip, which represents the path to follow, this path is modified as the obstacles move in the strips: also in this method we need a few of control points to perform the task.
\end{itemize}

Both methods were developed using  few discrete sensors: laser scanners, radars, cameras. These kind of sensors and the related algorithms are able to pursue  the task in a conservative way: there is a lack of information obviously due to the sensors  used and also because the  algorithms are based on  too few information, each body of the robot is described only from a few control points.

In addition some \emph{requirements} that an obstacle avoidance algorithm has to satisfy can be identified. They can be established after a careful analysis of some cases.

It is intuitive that the avoidance of obstacles needs some sensors which have a  field of view as large as possible.

 A robot equipped with a single camera or a single laser scanner can have information of the surrounding things only if these are in the sensor's field of view, what does it happen if something hits the robot from behind? The safety is guaranteed only for the parts of the robot close to sensor locations, while some parts of the robot are not because no proximity information is available.

Furthermore, to pursue the avoidance in real-time and in daily contexts we do not need a structured environment, a robot has to perform the avoidance everywhere: from the inside of an house to the street, in an airport or in a train station, in a market or in a farm. It needs no help from outside, just its sensor system has to help it. The robot has only the knowledge received from the sensors mounted on it, it should know only the path to follow and it should not know the obstacle position in advance.

The robot has to execute the avoidance as a task called only when it is needed, when there is not any obstacle, the robot has to follow its path. The change between the two tasks has to be autonomous: with its own intelligence the robot changes the task to afford.  

The requirements to be fulfilled can be summarized as follows :

\begin{itemize}
 \item \emph{Safety}: the robot performs the obstacle avoidance  in a secure way,  hitting no one;
 \item \emph{Availability}: the robot is always available to perform this task, in every possible environment condition;
 \item \emph{Real-Time}: the obstacle avoidance is done in a very short time, the time required to permit a prompt avoidance;
 \item \emph{Distributed Sensing}: the sensors are situated along all the surface of the robot, and we need only proximity information detected from the sensors mounted on the robot, anymore information situated in the environment is redundant;
\item \emph{Autonomous system}: the robot has autonomously change the priority of the tasks to perform.
\end{itemize}

\section{The distributed perception approach}

The requirements of \emph{Safety} and of \emph{Availability} describe an attitude to the protection of the robot. Recalling the Asimov's Laws, the robot needs to be protected not only for its own safety but also for safety of objects or human beings sharing its workspace.

Our idea is to think a robot protected using \emph{N} parallel springs distributed along all the  surface of the robot, taking in account all the elastic energy associated to them. 

\begin{figure}[ht]
\centering%
\includegraphics[height=180pt]{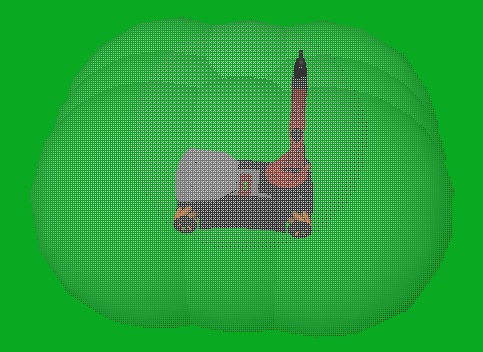}%
\caption{The springs protecting the robot \label{fig:energia1}}%
\end{figure}

As said, the robot is considered as \emph{protected} by N springs with a certain rest length. As an example, in Fig. \ref{fig:energia1} the safety volumes of all the springs are represented and activated, even though each spring is active only in proximity of an obstacle. In fact, when an obstacle \emph{touches} one or more springs (in the sense that it comes close enough to the corresponding sensors), the total elastic energy associated to these springs increases. To perform the obstacle avoidance, the robot has to move away from the obstacle so as to minimize the total energy, i.e. the sum of the elastic energies
of all the springs touched by the obstacle. 

\begin{figure}[ht]
\centering%
\includegraphics[height=180pt]{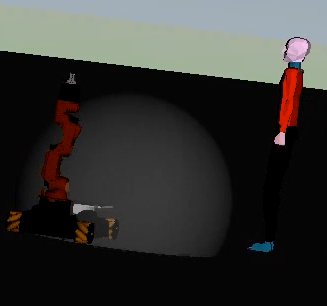}%
\caption{An example of avoidance using elastic energy \label{fig:energia}}%
\end{figure}

If we think of an anthropomorphic arm, to each body we should mount some proximity sensors . We imagine to associate an elastic energy to a sort of protective buffer centered in each sensor location. 
The elastic energy (figure \ref{fig:energia}) permits us to use distances  and the direction of the minimum distance as the only information to compute the algorithm. It's a distributed approach independent from the robot geometry.
The springs make up a protection area (figure \ref{fig:energia1}) where the control mechanisms are activated to perform the avoidance. The protection area is a rubber sphere where is defined an elastic energy. On each sensor there is this protection area.
On all its surface, the robot has these points of protection, we call them control points or sensor points. There is no more a single point of view, all the sensors distribute the perception along  the robot.  

The obstacle avoidance is obtained keeping constant to the initial value all the elastic energy associated to the sensors.

Since we use a proximity sensor skin (based on a finite number of N sensors distributed along the robot), the objective is : 

\begin{equation}
\sigma(q)=\sum_{k=1}^N \epsilon_k= K_i
\end{equation}

where $\epsilon_k$ is the elastic energy associated to the sensor point $k$ of the robot surface and $K_i$ an initial constant value.

\subsection{The proposed control algorithm}
We have used the NSB control to perform the avoidance in an autonomous way. 
\begin{figure}[ht]
\centering%
\includegraphics[height=280pt]{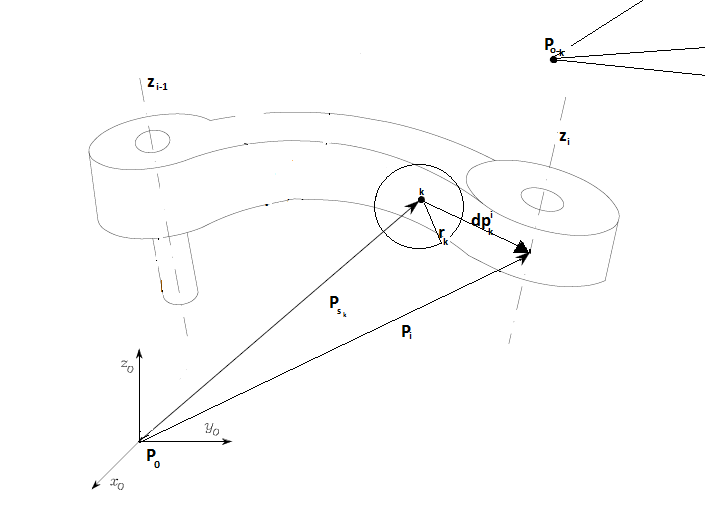}%
\caption{A generic arm with a sensor point and an obstacle. \label{fig:braccio}}%
\end{figure}%

Our method can be applied to a generic kinematic open chain. There is  only one hypothesis: the sensors provide to the control unit the point at minimum distance from the surface of the obstacle. Looking at the figure (\ref{fig:braccio}), in order to formalize the approach, assume that the $k$th sensor, located in the point $P_{s_k}$, measures the distance $d_k$ between
the nearest point on the obstacle $P_{o_k}$ and the sensor itself as
well as the direction of the minimum distance expressed by the unit
vector $v_{d_k}$. Assuming to adopt the Denavit-Hartenberg convention and
that all vectors expressed in frame 0 (the first fixed DH-frame of
the chain) have no superscript, with reference to
Fig.~\ref{fig:braccio}, let:
\begin{itemize}
    \item ${q}=[\begin{matrix} q_1&q_2&\cdots&q_n\end{matrix}]^T\in R^n$ be the vector of configuration variables;
 \item ${e_4}=[\begin{matrix} 0&0&0&1\end{matrix}]^T$;
	  \item $\tilde{p_{i-1}}={A_1^0(q_1)}\cdots A_{i-1}^{i-2}(q_{i-1}){e_4}$ be the position vector (homogeneous coordinates)
     of the origin of the DH-frame fixed to the link $i$;
    \item  ${z_{i-1}}={R_1^0(q_1)}\cdots {R_{i-1}^{i-2}(q_{i-1})}{z_0}$ be the $z$ axis (along joint $i$) of the frame fixed to link $i-1$;
    
    \item  ${p_{o_k}}$ be the position vector of the obstacle point at minimum distance from the $k$th sensor;
    \item  ${p_{s_k}}({q})={p_i({q})} - {R_i^0}({q}){d}{p_k^i}$ be the position vector of the $k$th sensor point, being $dp_k^i$ is the constant position of the $k$th sensor, assumed mounted on link $i$, with respect to frame $i$ fixed to link $i$.
\end{itemize}

The NSB control requires to define some task functions to perform, the behaviours. As described before our behaviours are two: the obstacle avoidance as primary behaviour and the path following as the second one. 

Our \emph{desired behaviour} is described by these equations, N is the number of sensors:

\begin{equation}
\sigma_d = 0; \dot{\sigma}_d=0; 
\end{equation}

\begin{equation}
\sigma(q)=\sum_{k=1}^N \epsilon_k
\end{equation}

The sigma function is the sum of the pseudo-energy of each \emph{k}th sensor. $q \in R^n$ is the configuration of the robot of $n$ degrees of freedom. 

\begin{figure}[ht]
\centering%
\includegraphics[height=280pt]{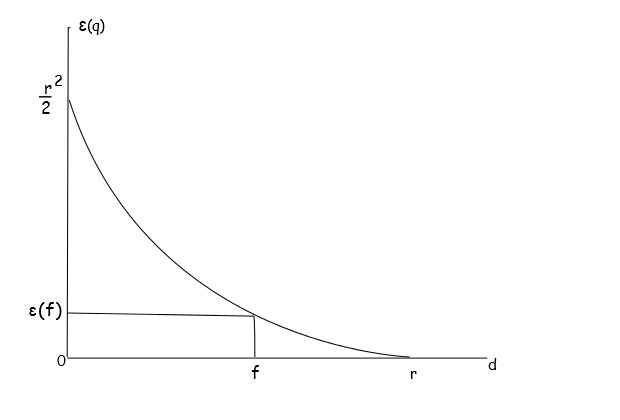}%
\caption{The pseudo-energy function. \label{fig:fenergia}}%
\end{figure}

The pseudo-energy (figure (\ref{fig:fenergia})) is a continuous function defined as follow:

\begin{equation}
\epsilon_k(q)=\left\{
\begin{split}
&\frac{1}{2}(d_k - r_k)^{2},\; \hbox{if} \; d_k \le r_k \\ &0, \; \hbox{otherwise} 
\end{split}
\right.
\end{equation}

It depends on the distance between the obstacle and the robot and the value $r_k \in R$, defined as the resting length of the spring. We define $f \in R$ as the activation threshold of the supervisor, $f_s \in R$ as the radius of the sensor's field of view and $d_k=\left\|P_{o_k}-P_{s_k}(q)\right\|$ the distance between  the position vector of the generic sensor point and the position vector of the obstacle point situated at minimum distance, both expressed in the base frame. The desired energy is zero. It means that we want to keep
\begin{equation}
d_k=r_k
\end{equation}
 when the obstacle avoidance behaviour is in active. 

We must satisfy these constraints:
\begin{equation}\label{e:sen}
f \le f_s;
\end{equation}
\begin{equation} \label{e:raggio}
f \le r_k;
\end{equation}
\begin{equation} \label{e:raggio1}
r_k \le f_s
\end{equation}

The equation (\ref{e:sen}) tell us that we can calculate the pseudo-energy only when the sensor can get information about the environment. The equation (\ref{e:raggio}) permits us to avoid making unnecessary calculations.
According to the equations (\ref{e:sen}),(\ref{e:raggio}), (\ref{e:raggio1}) we have to choose $r_k$ so that we have :
\begin{equation}
f \le r_k \le f_s
\end{equation}
When the obstacle avoidance task is activated,  we have an initial error (equation (\ref{e:init})) in the CLIK:
\begin{equation}\label{e:init}
e_0=\epsilon(f)
\end{equation}
In discrete time the value of $e_0$ can compromise the convergence of the CLIK algorithm (see the discussion in \cite{rif17}). 

Each task in the NSB control has to be computed in the CLIK algorithm, before to merge both behaviours using a task combination function.

In literature there is no a specific way to design the supervisor in the NSB control. We have elaborated a regularized supervisor: a convex combination of the different tasks, the \textit{task combination function}.  

The $\gamma$ gains play an important role in relations between the different behaviours, they may be seen as the elastic constants of springs that bring the robot on the tasks of different behaviours.

\subsection{The task combination function}

A velocity vector for each behaviour is computed as if it was acting alone; then, before adding the single contribution to the overall configuration velocity, a lower-priority behaviour is projected onto the null space of the higher-priority behaviours so as to remove those velocity components that would conflict with it.
A supervisor is used to change the priorities of the tasks. 

A simple way to do it, it is to use a finite state machine, changing from state to state in a discrete way. It means that \emph{the supervisor does not permit to merge the two tasks and gradually select one of them}:  there is only one behaviour acting at a time. This kind of supervisor can generate in the robot sudden accelerations.

We think the supervisor not as a crispy selector of behaviours but as a continuous  function that acts as a weight to the velocities of the different behaviours, i.e. as it is a convex combination of the different tasks.

Depending on the number of the tasks and the kind of application to develop you can choose different functions.

Our application uses two behaviours: one for the avoidance one for the path following. They are defined as:
\begin{equation}
\dot{q}_o=J_o^\dagger(q)[\dot{\sigma}_d+\gamma_o\tilde{\sigma}_d]+ (I-J_o^\dagger(q)J_o(q))\dot{q}_g
\end{equation}
\begin{equation}
\dot{q}_g=J_g^{\dagger}(\dot{x}_d+\gamma_g(x_d-x))
\end{equation}
where $q_o \in R^n$ is the configuration velocity vector due to the obstacle avoidance, $q_g \in R^n$ is the configuration velocity vector due to the achievement of the goal position, $(x_d, \dot{x_d}) \in R^M$ is the desired position and velocity vector expressed in the base frame, $J_g$ is a $M\times n$ matrix.
We defined the following continuous function:
\begin{equation}
\lambda(d) = \frac{1}{\pi}\arctan (-K(d-f))+\frac{1}{2}
\end{equation}
where 
\begin{equation}
d= \min_k(d_k)
\end{equation}
where $d_k=\left\|P_{o_k}-P_{s_k}\right\|$ and \emph{f} is a threshold,we have to declare  $f < r_k$ to obtain an initial energy different from zero, as we said before. It is the value where the function has its mid-scale value. 

Note that in this way a sort of ``grey zone'' is defined where a
priority among the tasks is not ``crisply'' defined. Outside of this
zone (namely for $d$ large or small enough), the priority is
established by the classical null-space projection method.
Varying K varies the slope of the function. 

\begin{figure}[ht]
\centering%
\includegraphics[height=140pt]{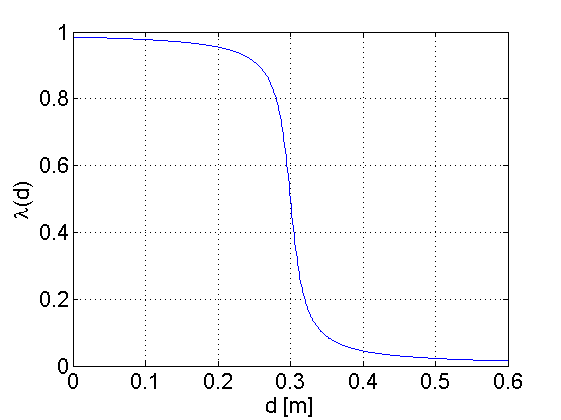}%
\caption{An example of the task combination function. \label{fig:supervisore}}%
\end{figure}%

The figure (\ref{fig:supervisore}) shows an example of the function.

\begin{figure}[ht]
\centering%
\includegraphics[height=180pt]{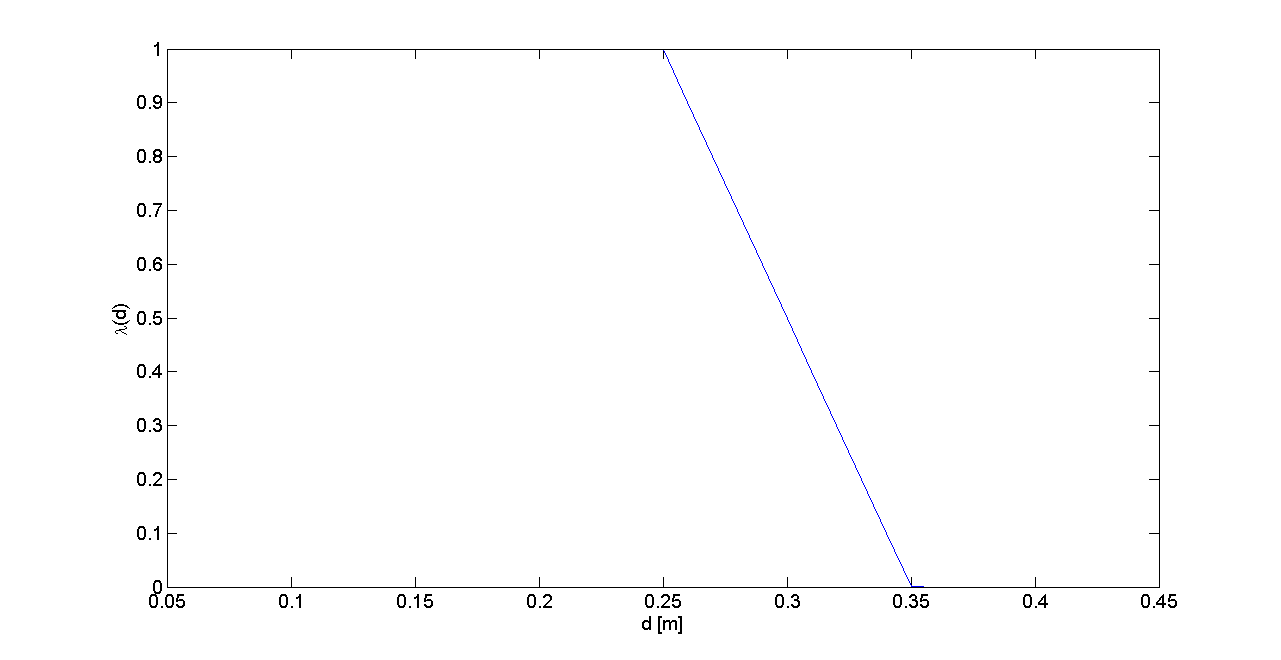}%
\caption{An example of the piece-wise linear task combination function. \label{fig:supervisore1}}%
\end{figure}

We can use also a piece-wise linear function (figure (\ref{fig:supervisore1})) defined as follows:

\begin{equation}\label{e:linear}
\lambda(d)=\left\{
\begin{split}
&-\frac{(d-f-\epsilon)}{2\epsilon},\;\hbox{if}\; (f-\epsilon)\le d \le (f+\epsilon) \\ &1, \; \hbox{if}\; d<(f-\epsilon) \\ &0,\; if \;d>(f+\epsilon)
\end{split}
\right.
\end{equation}
where 
\begin{equation}
d= \min_k(\left\|P_{o_k}-P_{s_k}\right\|)
\end{equation}

The system velocity is then defined as the convex combination:
\begin{equation}\label{e:comb}
\dot{q}= \lambda(d)\Bigg(J_o^\dagger(q)[\dot{\sigma}_d+\gamma_o\tilde{\sigma}_d]+ (I-J_o^\dagger(q)J_o(q))\dot{q}_g\Bigg)+(1-\lambda(d))(\dot{q}_g) 
\end{equation}
where:
\begin{equation}
\tilde{\sigma_d}=\sigma_d-\sigma(q);
\end{equation}
\begin{equation}
\dot{q_g}=J_g^{\dagger}(\dot{x}_d+\gamma_g(x_d-x));
\end{equation}

When the minimum distance between the obstacle and our robot is less then the threshold our supervisor gives more importance to the obstacle avoidance behaviour, in this case $\lambda(d)$ goes to $1$ as much as \emph{d} is less then \emph{f}; otherwise $\lambda(d)$ goes to $0$.

\subsection{The Jacobians $J_{o}$ and $J_g$}

$J_{o}$ is the one associated to the $\sigma$ function and it is the gradient of the elastic energy found before. It depends on the position of the sensor, the configuration of the robot and the minimum distance from the obstacle. We have calculated it in a closed form: 

\begin{equation}
J_{o}(q)=\frac{\partial\sigma(q)}{\partial{q}}=\sum_{k=1}^N\frac{\partial\epsilon_k(q)}{\partial{q}};
\end{equation}

Defined $d_k=\left\|P_{o_k}-P_{s_k}\right\|$ and the unit vector $v_k=\frac{(P_{o_k}-P_{s_k})}{\left\|P_{o_k}-P_{s_k}\right\|}$ we have:
\begin{equation}
\frac{\partial\epsilon_k(q)}{\partial{q}}=
\left\{
\begin{split}
-(&d_k-r_k)v_k^T\frac{\partial{P_{s_k}(q)}}{\partial{q}},& \hbox{if} \;d_k \le r_k \\ &0, \; \hbox{otherwise}
\end{split}
\right.
\end{equation}
and $\frac{\partial{P_{s_k}(q)}}{\partial{q}}$ is the Jacobian for the \textit{k}th sensor.

Instead $J_g$ is the geometric Jacobian associated to the configuration $q$ of the robot.

\subsection{The architecture of the control unit}

The control algorithm described before can be developed in more ways, we suggest an architecture.
The control has to respect the system velocity in \ref{e:comb}.

 The design consists of different modules:
\begin{itemize}
    \item the sensors system, where the proximity information are elaborated;
	 \item the $\sigma$ unit, that calculates the energy for each sensor;
	 \item the two units that compute the jacobians for the CLIK
implementation;
	\item the $k(q)$ unit that calculates the direct kinematics function
of the robot
    \item the task combination function unit.
\end{itemize}

\begin{figure}[ht]
\centering%
\includegraphics[height=200pt]{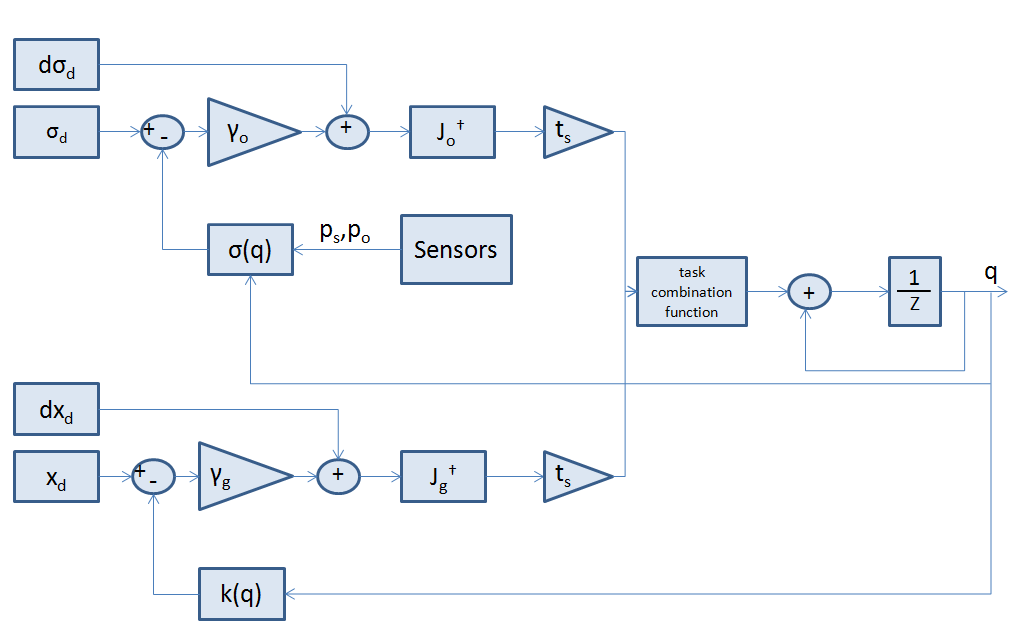}%
\caption{The control scheme. \label{fig:schema}}%
\end{figure}

The sensors system elaborates the information coming from the sensors and it sends them to the $\sigma$ and $J_o$ units and to the supervisor one.
The figure (\ref{fig:schema}) shows a scheme of the system. 

The control unit needs the path ($x_d,\dot{x_d}$) generated from a planning unit. In our idea the planning algorithm does its work separately from the control. Note that $t_s$ is the sampling time and the gains
$\gamma_g$ and $\gamma_o$ are strictly related to its value.
  
\chapter{The Kuka Youbot Simulator}\label{c:quarto}

To evaluate the new algorithm we developed a dynamical simulator.  The robot simulated is the Kuka Youbot a mobile robot with 8 degrees of freedom. The first three are for the base, two prismatic joints and a revolute one. The last fives are for the arm mounted on its base. It's a small new robot useful for doing research. 

The simulator is developed using two toolboxes of Simulink (Matlab) : SimMechanics and VR ToolBox.

\section{SimMechanics and the VR ToolBox}

SimMechanics is a toolbox that extends the  Simulink capabilities. It allows the modelling and simulation of mechanical systems. It is based on the principle to be able to model a mechanical system without the knowledge of the dynamic equations associated to it. It uses blocks that represent bodies, joints, constraints and drivers. In order to model a mechanical system we must know only the structure of the various entities.

 An example would be the classic model of an anthropomorphic arm: it consists of a base on which are applied two revolute joints  having axes perpendicular to each other, one of these is connected to the first arm of the manipulator, which then is connected to a second arm through another joint. For the realization of the model of the arm we need three bodies (the base and the two arms) and three revolute joints.

For the modelling of the bodies we used a block of the library called Body, in it  these fields are inserted:
\begin{itemize}
    \item mass;
     \item moment of inertia;
     \item position and rotations of the CS (Coordinate System,a frame that can be connected to a point of the block);
    \item position and rotation of the CG (Center of Gravity, the frame of the Center of Gravity of the body).
\end{itemize}
We can found in the SimMechanics library  the following blocks:
\begin{itemize}
    \item Ground, to model the earth;
    \item Machine Environment, to define the properties of the environment in which you are working, for example, the definition of the gravity vector;
	 \item Shared Machine Environment,to share the same properties of the environment  among multiple objects that do not belong to the same kinematic chain.
\end{itemize}

Therefore  by simple connections between blocks SimMechanics allows the realization of a simulator. For each CS is possible to connect the sensors to read the information of: position, velocity, angular velocity, rotation matrix, acceleration, angular acceleration. In addition to the sensors in the library there are the actuators for both the joints and the bodies: the actuators permit us to control joints and bodies in position and in terms of forces.

The Virtual Reality Toolbox allows the use of a virtual reality viewer in Simulink. The 3D environments used in the VR Toolbox must be developed through language VRML97. Many CAD drawing softwares are enabled to export in this format. For this work we have generated the VRML model of the robot starting from the one made by Kuka for the open source software Blender. 

We must add that we took two conversions to obtain the final VRML model:  one  from the software Blender to 3D Studio Max, and another one from the latter to the VRML editing software.

Each object has several properties, among which the most significant are:
\begin{itemize}
	\item Center, it is the vector that indicates the point at which the center of the object is fixed in the space;
	\item Rotation, it is the vector and the angle indicating the rotation angle of the object in  axis-angle representation;
	\item Translation, it is the vector that indicates the translational motion of the object relative to its initial position;
	\item Shape, it is a property divided into Appearance and Geometry in which is expressed the geometry and the colours of the object.
\end{itemize}
These properties can be changed when the object is inserted in a simulator using Simulink block \textit{VRsink}. In it you can associate a 3D scene to a simulator and you can modify the properties of the same 3D scene with the classic Simulink signals.

\subsection{The Kuka Youbot robot} 
The Kuka Youbot is a robot formed by a omnidirectional base and a manipulator with five degrees of freedom at the end of which is placed a gripper to grasp objects.

 Its mechanical structure is represented by an open kinematic chain, with 8 degrees of freedom: 3 for the base and 5 for the manipulator.
\begin{figure}[ht]
\centering%
\includegraphics[height=240pt]{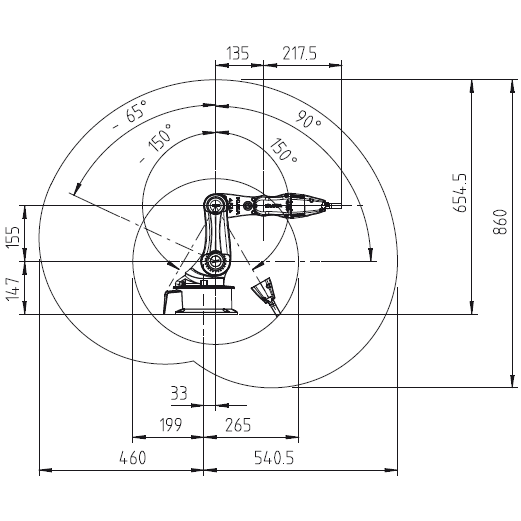}%
\caption{Description of the Youbot \label{fig:youbot1}}%
\end{figure}%

The omnidirectional base can be represented with two prismatic and one revolute joints, the manipulator with five revolute joints.

The base is made of steel and  it weighs 20 kg, instead the arm is made of an alloy of aluminium and magnesium. The arm supports 0.5 kg of payload. It presents a repeatability of 1 mm. 
We report a list of the kinematics and dynamics information of the Youbot.
In the list below each rigid body has a relative position and orientation with respect to its parent frame. Orientation is described in Euler angles. The composite rotation convention is Yaw Pitch Roll. Joint axis is the axis of rotation for revolute joint or the axis of translation for prismatic joint. Joint axis is specified in the joint frame. The dynamics related parameters are mass, relative pose of center of mass and moment of inertia tensor. 

\begin{itemize}
\item \emph{Base Frame}: Relative positionXYZ="0mm 0mm 84mm", Mass=19.803 Kg;

\item \emph{front-left wheel joint}: Parent Joint = "base frame", Relative positionXYZ = "228mm 158mm -34mm", Joint type = "continuous", Joint axis= "0 1 0", Mass = "1.40kg";
\item \emph{front-right wheel joint}: Parent joint="base frame",Relative positionXYZ = "228mm -158mm -34mm",Joint type = "continuous",Joint axis = "0 1 0", Mass = "1.40kg";
\item \emph{back-left wheel joint}: Parent joint = "base frame",Relative positionXYZ = "-228mm 158mm -34mm", Joint type = "continuous", Joint axis = "0 1 0", Mass = "1.40kg";
\item \emph{back-right wheel joint}: Parent joint = "base frame", Relative positionXYZ = "-228mm -158mm -34mm", Joint type = "continuous", Joint axis = "0 1 0", Mass = "1.40kg";
\item \emph{plate}: Parent joint = "base frame", Relative positionXYZ = "-159mm 0mm 46mm", Joint type = "fixed", Mass = "2.397kg";
\item \emph{arm base frame}: Parent joint = "base", Relative positionXYZ = "143mm 0mm 46mm", Joint type = "fixed", Mass = "0.961kg";
\item \emph{arm joint 1}: Parent joint = "arm base frame", Relative positionXYZ = "24mm 0mm 115mm", OrientationZYX = "0° 0° 180°", Joint type = "revolute", Joint axis = "0 0 1", Joint limits = "-169° 169°", Mass = "1.390kg", max. torque = "9.5Nm", principal axis of inertia: positionXYZ = "15.16mm 3.59mm 31.05mm", orientationZYX = "180° 20° 0°", inertia = "xx = 0.0029525 $kg\;m^2$,yy=0.0060091 $kg\;m^2$,zz=0.0058821 $kg\;m^2$";

\item \emph{arm joint 2}: Parent joint = "arm joint 1", Relative positionXYZ = "33mm 0mm 0mm", Orientation ZYX = "-90° 0° 90°", Joint type = "revolute", Joint axis = "0 0 1", Joint limits = "-65° 90°", Mass = "1.318kg", max. torque = "9.5Nm", principal axis of inertia:positionXYZ = "113.97mm 15.0mm -19.03mm",    orientationZYX = "-90° 0° -90°", inertia = "xx = 0.0031145$kg\;m^2$ yy = 0.0005843$kg\;m^2$ zz=0.0031631$kg\;m^2$".

\item \emph{arm joint 3}: Parent joint = "arm joint 2", Relative positionXYZ ="155mm 0mm 0mm", OrientationZYX = "-90° 0° 0°", Joint type = "revolute", Joint axis = "0 0 1", Joint limits = "-151° 146°", Mass = "0.821kg", max. torque = "6.0Nm", principal axis of inertia: positionXYZ = "0.13mm 104.41mm 20.22mm",    orientationZYX = "0° 0° 90°",inertia = "xx = 0.00172767$kg\;m^2$ yy = 0.00041967$kg\;m^2$ zz=0.0018468$kg\;m^2$".

\item \emph{arm joint 4}: Parent joint = "arm joint 3", Relative positionXYZ = "0mm 135mm 0mm", OrientationZYX = "0° 0° 0°", Joint type = "revolute", Joint axis = "0 0 1", Joint limits = "-102.5° 102.5°", Mass = "0.769kg", max. torque = "2.0Nm"; principal axis of inertia: positionXYZ = "0.15mm 53.53mm -24.64mm", orientationZYX = "0° 180° 40°",inertia = "xx = 0.0006764$kg\;m^2$ yy = 0.0010573$kg\;m^2$ zz=0.0006610$kg\;m^2$".

\item \emph{arm joint 5}: Parent joint = "arm joint 4", Relative positionXYZ = "0mm 113.6mm 0mm"; OrientationZYX = "0° 0° -90°", Joint type = "revolute", Joint axis = "0 0 -1", Joint limits = "-165° 165°", Mass = "0.687kg", max. torque = "1.0Nm", principal axis of inertia: positionXYZ = "0mm 1.2mm -16.48mm", orientationZYX = "0°90°0°"; inertia = "xx = 0.0001934$kg\;m^2$ yy = 0.0001602$kg\;m^2$ zz=0.0000689$kg\;m^2$ ";

\item \emph{gripper base frame}: Parent joint = "arm joint 5", Relative position ="0mm 0mm 57.16mm", OrientationZYX = "180° 0° 0°", Joint type = "fixed", Mass = "0.199kg", principal axis of inertia: positionXYZ = "0mm 0mm 28.9mm",   orientationZYX = "180° 0° 90°", inertia = "xx = 0.0002324$kg\;m^2$ yy = 0.0003629$kg\;m^2$ zz=0.0002067$kg\;m^2$".

\item \emph{gripper left finger joint}: Parent joint = "gripper base frame", Relative position = "0mm 8.2mm 0mm", Joint type = "prismatic", Joint axis = "0 1 0", Joint limits = "0mm 12.5mm", Mass = "0.010kg".

\item \emph{gripper right finger joint}: Parent joint = "gripper base frame", Relative position = "0mm -8.2mm 0mm", Joint type = "prismatic", Joint axis = "0 -1 0", Joint limits = "0mm 12.5mm", Mass = "0.010kg".

\end{itemize}

For the computation of the Jacobian we have find out also the Denavit-Hartenberg parameters of the Youbot. They are:
\begin{center}
\begin{tabular}{l*{6}{c}r}
Joint & $\theta[rad]$ & $d[m]$ &$\alpha[rad]$ &$a[m]$ \\
\hline
1  			&0 &q1 &$\frac{\pi}{2}$ &0 \\
2           &$\frac{\pi}{2}$ &q2 &$\frac{\pi}{2}$ &0  \\
3           &q3 &0 &0 &0.167 \\
4           &q4 &0.147 &$\frac{\pi}{2}$ &0.033  \\
5           &q5 &0 &0 &0.155 \\
6           &q6 &0 &0 &0.135 \\
7           &q7 &0 &$\frac{\pi}{2}$ &0 \\
8           &q8 &0.2175 &0 &0 \\
\end{tabular}    
\end{center}

\section{The Simulator}

The realized simulator  consists of two macro blocks: one models the mobile base and the other one  implements the manipulator.

The model of the base has been achieved with the connection of a body (which has been described with the moment of inertia and the geometry of the object) to a prismatic joint with two degrees of freedom (\textit {respect to the axis x and y axis}) and to a revolute joint {to implement a rotation about the \textit{z axis}}.

The model of the arm consists of 5 blocks, one for each joint-body connection. For every body that makes up the kinematic chain was transcribed the moment of inertia, mass, and CS for each point where you can connect a sensor useful for finding the needed information.

\begin{figure}[ht]
\centering%
\includegraphics[height=200pt]{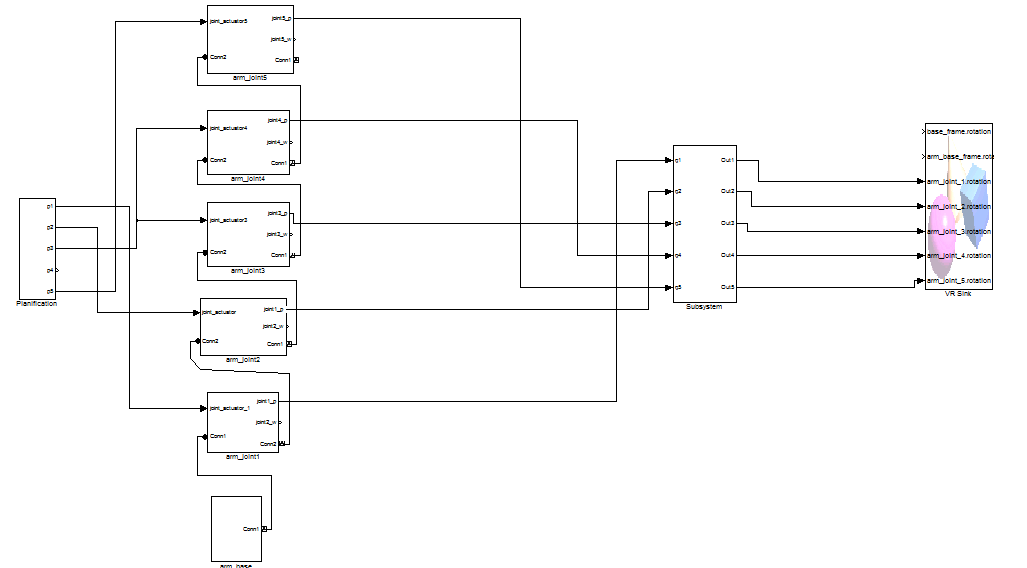}%
\caption{The Simulator. \label{fig:sim}}%
\end{figure}%

In the simulator is also included the VR Sink for visualization of the 3D model in real time. It has been connected to the measurements made by the sensors for the movement of the joints and for the movements of the machine.

\begin{figure}[ht]
\centering%
\includegraphics[height=150pt]{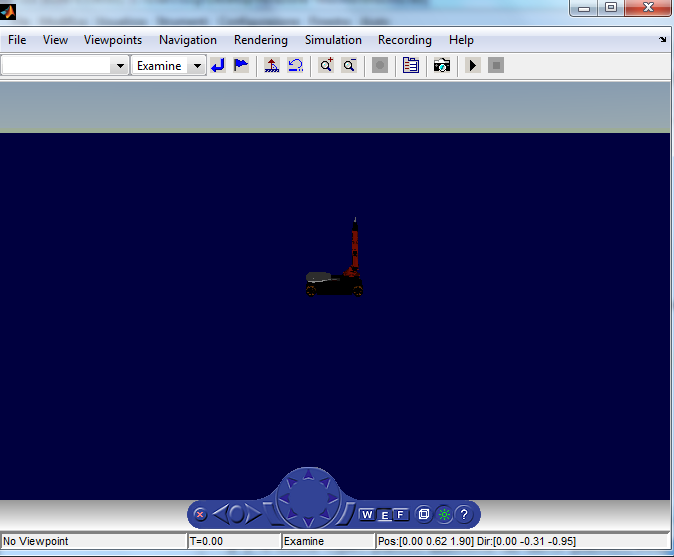}%
\caption{ 3D scene \label{fig:sim3D}}%
\end{figure}%

The 3D environment realized currently includes the presence of:

\begin{itemize}
    \item \textit{Youbot Kuka}, it is the robot model;
	 \item \textit{plane}, it is a 3D model of a floor, on which the robot is put;
	 \item \textit{object}, is a generic object situated in the scene.
\end{itemize}

\section{Description of rigid bodies, actuators and sensors}

\begin{figure}[ht]
\centering%
\includegraphics[height=150pt]{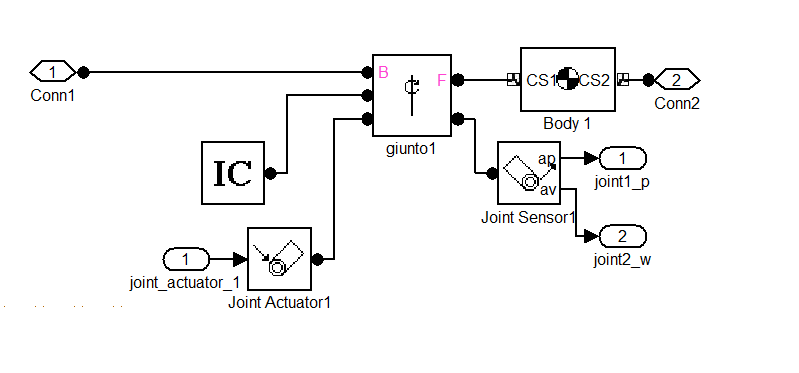}%
\caption{The model of a single body \label{fig:corpo}}%
\end{figure}

\begin{figure}[ht]
\centering%
\includegraphics[height=250pt]{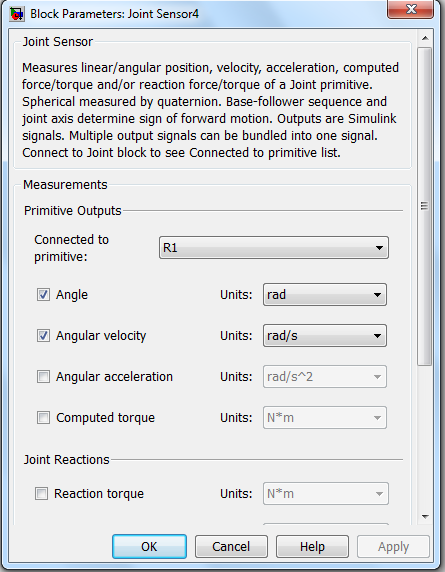}%
\caption{Window for the sensor \label{fig:sensore}}%
\end{figure}

\begin{figure}[ht]
\centering%
\includegraphics[height=250pt]{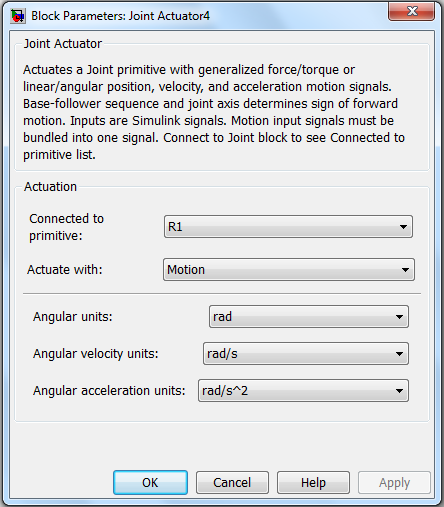}%
\caption{Window for the actuator\label{fig:attuatore}}%
\end{figure}

Each rigid body within the simulator has the following components, presented in \figurename~\ref{fig:corpo}:
\begin{itemize}
    \item \textit{Joint}, it is the joint to which the body is connected,for the manipulator we used five revolute joints, for the base two prismatic and one revolute joints;
	 \item \textit{IC}, the initial conditions in which the joint is placed;
	 \item \textit{Body}, it is the body linked to the joint;
	 \item \textit{Joint Sensor},(in \figurename~\ref{fig:sensore}) it is the sensor connected to the joint, it permits to obtain different information (angle, angular velocity, angular acceleration, torque for the revolute joints; position, velocity and acceleration for prismatic joints);
	 \item \textit{Joint Actuator},(in \figurename~\ref{fig:attuatore} ) it is the actuator that allows the application of the signals in terms of position or force (for prismatic joints), or in angle of rotation and torque (for revolute joints);
\end{itemize}

The realized simulator provides a path planned in joint space. It is possible to make a planning in terms of forces simply by changing the properties of the Joint Actuators. In position a Joint Actuator requires three signals indicating: position, velocity and acceleration. Obviously, these three quantities are angular when it comes to a revolute joint. In the case where it is desired to control the robot in terms of torque then it is possible to associate signals in forces.

\begin{figure}[ht]
\centering%
\includegraphics[height=250pt]{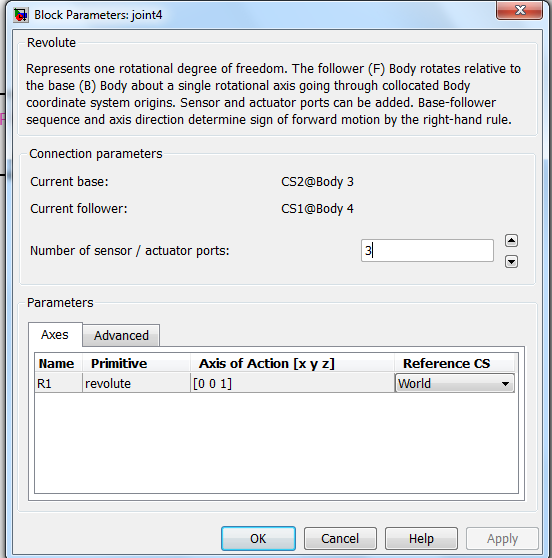}%
\caption{Window for the joint \label{fig:giunto}}%
\end{figure}

For each joint must be given the axis around which rotation occurs (\textit{Axis of Action}, whose coordinates can be related to Frame World, or to the frame of the body that precedes or follows the same joint) or if it is prismatic the vector on which the translation takes place.  In the block Joint (in \figurename~\ref{fig:giunto}) it also indicates the number of sensors and actuators connected to it.

Each introduced body (in \figurename~\ref{fig:body})  must be connected to a joint and we have to specify:
\begin{figure}[ht]
\centering%
\includegraphics[height=250pt]{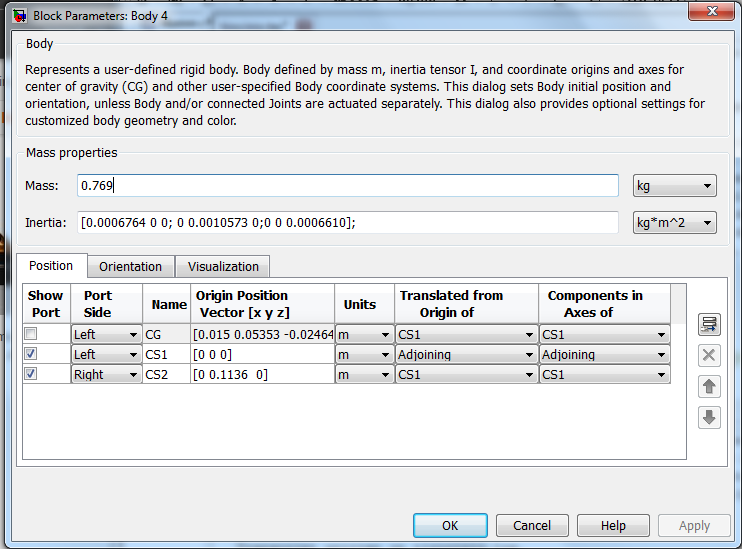}%
\caption{Window to modify the body\label{fig:body}}%
\end{figure}%
\begin{itemize}
    \item the \textit{mass} expressed in  specified units;
	 \item the \textit{inertia};
	 \item the \textit{CG} and the \textit{CS}, the first defined as the point in which  the center of mass of the body is situated, the latter points in which it is possible to apply other bodies, joints, sensors or actuators. For the realization of the simulator they have been used substantially to connect the prior joint  to the next one, taking into account of course the exact position expressed in the manufacturer's specifications.

\end{itemize}
\subsection{How SimMechanics analyses the motion}

The Parameters tab of the Machine Environment dialog allows you to choose the analysis mode you want to simulate in. You make this choice via the Analysis mode pull-down menu. In the case of linearization, use the Linearization tab to set the size of the small perturbations. See the Analyzing Motion chapter for detailed instructions and examples concerning the motion analysis modes.

By choosing one of these analysis modes, you implement the type of motion analysis you want. 
Here there is a list of the analysis modes offered:
\begin{itemize}
    \item \textit{Forward dynamics}, computes the positions and velocities of a system's bodies at each time step, given the initial positions and velocities of its bodies and any forces applied to the system;
	\item	 \textit{Linearization}, computes the effect of small perturbations on system motion through the Simulinklinmod command;
	\item \textit{Trimming}, enables the Simulinktrim command to compute steady-state solutions of system motion;
	\item \textit{Inverse dynamics (open-loop)}, computes the forces required to produce a specified velocity for each body of an open-loop system;
	\item \textit{Inverse dynamics (closed-loop)}, computes the forces required to produce a specified velocity for each body of a closed-loop machine.
\end{itemize}

SimMechanics uses an ODE solver to solve the system's equations of motion, typically in tandem with a constraint solver. Simulink provides an extensive set of ODE solvers that represent the most advanced numerical techniques available for solving differential equations in general and equations of motion in particular. The Solver pane of a model's Simulation Parameters dialog box allows you to select any of these solvers for use by Simulink in solving the model's dynamics.

The Dormand-Prince solver (ode45) that Simulink uses by default works well for many mechanical systems, but might require too much time to solve systems that are stiff, that is, have bodies that move at widely varying speeds or that have many discontinuities in their motion. An example of a stiff system is a pair of coupled oscillators in which one oscillator is much lighter than the other and hence oscillates much more rapidly. Any of the following solvers might require significantly less time than the default solver to solve a stiff system:
\begin{itemize}
    \item \textit{ode15s}, variable-order solver based on a backward differentiation rule (variant of Gear's method); 
	 \item \textit{ode23t}, trapezoidal rule solver. Use this solver if your system is slightly stiff, to avoid numerical damping;
	 \item \textit{ode23tb}, implicit Runge-Kutta method solver combining trapezoidal rule and a backward differentiation rule of order 2. More efficient than ode15s if the solution has many discontinuities;

	 \item \textit{ode23s}, Modified Rosenbrock method solver of order 2. This solver is also more efficient than ode15s, if the solution has many discontinuities. 
\end{itemize}
In the developed simulator we used the \textit{ode23s} solver with Forward dynamics analysis.
\section{Connecting to the Virtual Reality tools}

\begin{figure}[ht]
\centering%
\includegraphics[height=150pt]{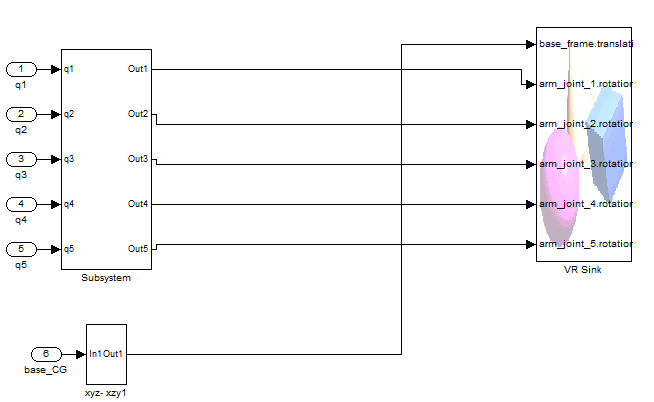}%
\caption{ 3D visualization \label{fig:visual}}%
\end{figure}

To connect to the VR Sink the signals processed by the simulator developed we must treat them (in \figurename~\ref{fig:visual} ), we need to apply a rotation from world-frame of the simulator to the 3D environment world-frame.

\begin{figure}[ht]
\centering%
\includegraphics[height=250pt]{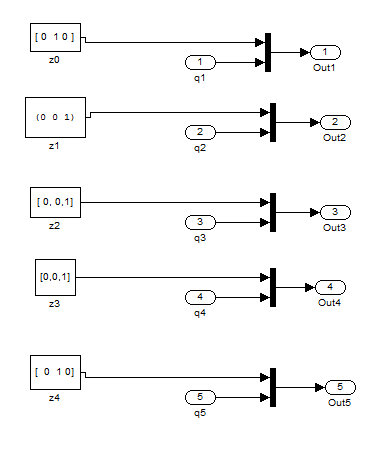}%
\caption{Signals for the rotations.\label{fig:assi}}%
\end{figure}

 The objects in the environment can change the properties of the signals through Simulink. Among these properties we mention the most relevant, used in the simulator: rotations and translations. The signals for the rotations are introduced (in \figurename~\ref{fig:assi} ) as a vector of four elements: indeed, they follow an axis-angle representation, three elements for the axis and one for the rotation in radians. For the translation we need a vector, it contains the position coordinates (opportunely shifted if in the 3D world we have a different world frame from the one used in SimMechanics).

\begin{figure}[ht]
\centering%
\includegraphics[height=150pt]{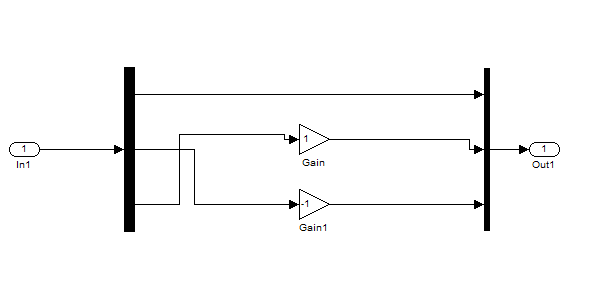}%
\caption{components for  the translations \label{fig:cambio}}%
\end{figure}

We have identified a substantial difference between the coordinate system (in \figurename~\ref{fig:world} ) of SimMechanics with the one of the VR Builder:
\begin{figure}[ht]
\centering%
\includegraphics[height=150pt]{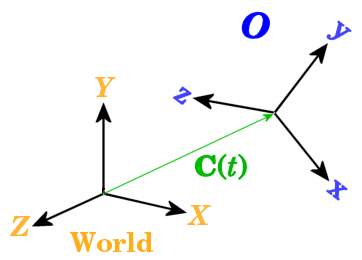}%
\caption{Axis World in SimMechanics \label{fig:world}}%
\end{figure}%
\begin{itemize}
    \item \textit{axis x of SimMechanis}, it is the same in the two environments;
	 \item \textit{axis y of SimMechanis}, it corresponds in the virtual  \textit{(-z)-axis};
 	\item \textit{axis z of SimMechanics}, it corresponds to the \textit{y axis} in the environment. 
\end{itemize} 
There was therefore a necessary correction of the axis expressed in rotations and  of the components used in translations (in \figurename~\ref{fig:cambio} ).

\subsection{VRML: the Virtual Reality Modeling Language. }
The 3D software used to model the robot and the environment is the one provided by Matlab: Logos VR Builder . You can make a 3D scene using the instruments given by the software or by writing code.

 The language to use is the VRML (Virtual Reality Modeling Language). It is a standard file format for representing 3D interactive vector graphics, designed particularly for the World Wide Web. VRML is a text file format where, e.g., vertices and edges for a 3D polygon can be specified along with the surface colour, UV mapped textures, shininess, transparency, and so on. 

It has a format similar to XML. Theoretically, the objects can contain anything (3D geometry, MIDI data, JPEG images, etc..). VRML defines a set of objects useful for doing 3D graphics. These objects are called Nodes.

Nodes are arranged in hierarchical structures called scene graphs. A node has the following characteristics:
\begin{itemize}
    \item \emph{What kind of object it is;}
	 \item \emph{The parameters that distinguish this node from other nodes of the same type;}
	 \item \emph{A name to identify this node;}
	 \item \emph{Child nodes.}
\end{itemize}

It offers basic primitives to make up a scene:

\begin{itemize}
    	\item \emph{Sphere;} 
		\item \emph{Cone;}
	   \item \emph{Cube;}
      \item \emph{Triangle;}
      \item \emph{Box;}
      \item \emph{Cylinder;}
\item \emph{ElevationGrid;}
\item \emph{Extrusion;}
\item \emph{IdexedFaceSet;}
\item \emph{IndexedLineSet;}
\item \emph{PointSet;}
\item \emph{Text;}

\end{itemize}

The Box, Cone, Cylinder, and Sphere are geometric primitives. The Text node displays a string with a specified font style. The WorldInfo node holds the world's title and other information, such as author and copyright. The ElevationGrid node creates surfaces and terrains. The Extrusion node creates solids by sweeping a 2D cross-section though a 3D spine. The IndexedFaceSet, IndexedLineSet, and PointSet nodes use Coordinate nodes to create solid faces, lines, and points, respectively. These raw geometry nodes give you more flexibility than the geometric primitives, and can actually create more efficient VRML worlds.

Each node can be inserted in a group. There is a hierarchical structure of the scene. Each node of the Kuka Youbot model has to respect the hierarchy of the joints of the robot.

There are also some basics camera and light features. We can insert in the scene the following types of light:
\begin{itemize}
    \item \textit{point light};
\item \textit{spot light};
\item \textit{directional light}.
\end{itemize}
We can define geometrical properties of the shape such as the ones for the material:
\begin{itemize}
    \item \textit{ambientIntensity};
	\item \textit{diffuseColor};
	\item \textit{emissiveColor};
	\item \textit{shininess};
   \item \textit{specularColor};
   \item \textit{trasparency}.
\end{itemize}
You can control the diffuse (shading) color, emissive (glow) color, transparency, shininess, and the other optical properties of an object using its Appearance node's Material field. These optical properties interact with the scene lighting to determine the image presented to the viewer. 

Only some kinds of materials are well described by the optical properties specified in the Material field. Metal and glass are; wood, tile, and painted objects are not. You can override optical properties with Color nodes, or wrap textures around the 3D shapes. VRML supports three kinds of texture mapping fields: \textit{ImageTexture} (from JPEG, PNG, and GIF files), \textit{PixelTexture} (from raw image data), and \textit{MovieTexture} (from MPEG1 files). If you use an image with an alpha channel for texture, you can create "holes" in the texture mapping that allow the underlying material properties to show through.
  
Ground color, sky color, and background textures are defined by the \textit{Background node}. Multiple backgrounds can be kept in a stack and bound dynamically. Atmosphere and an increased sense of depth can be created by using a \textit{Fog node}. You can provide \textit{Viewpoint nodes} to help the user navigate your world.

VRML files measure distance in meters, angles in radians, time in seconds, and colours as RGB triplets with each value in the range [0.0,1.0]. It uses a Cartesian, right-handed three-dimensional coordinate system. By default, the viewer is positioned along the positive Z-axis so as to look along the -Z direction with +Y-axis up. The Translation and Rotations have these representations : 
\begin{itemize}
    \item translation: single vector with three elements;
	\item rotation: a vector of four element, for the axis-angle representation.
\end{itemize}

\section{Obstacle modelling}
In SimMechanics it is simple to model obstacles, they are represented by a body block, where we have to specify the moment of inertia tensor, the mass and obviously the position in the space where the obstacle is placed.
We report here some moment of inertia tensor used to model obstacles. The first one is a solid sphere of radius $r$ and mass $m$:
$$
I_{sphere} =\left[
\begin{array}{ccc}
\frac{2}{5}mr^2 & 0 & 0 \\
0 & \frac{2}{5}mr^2 & 0 \\
0 & 0 &  \frac{2}{5}mr^2
\end{array}
\right]
$$

We used also a solid cuboid of width $w$, height $h$, depth $d$, and mass $m$ and a solid cylinder of radius $r_c$, height $h_c$ and mass $m_c$:
$$
I_{cuboid} =\left[
\begin{array}{ccc}
\frac{1}{12}m(h^2 +d^2) & 0 & 0 \\
0 & \frac{1}{12}m(w^2+d^2) & 0 \\
0 & 0 &  \frac{1}{12}m(w^2+h^2)
\end{array}
\right]
$$

$$
I_{cylinder} =\left[
\begin{array}{ccc}
\frac{1}{12}m_c(3r_c^2 +h_c^2) & 0 & 0 \\
0 & \frac{1}{12}m_c(3r_c^2+h_c^2) & 0 \\
0 & 0 &  \frac{1}{12}m_c(r_c^2)
\end{array}
\right]
$$

In some cases we used the analytic geometry equations of different elements. For example we mention the equation of a sphere with center in $(x_o,y_o,z_o)$ and radius $r$. The points on the sphere with radius $r$ can be parametrised with following equations:
\begin{eqnarray}
x=x_o+r cos(\theta)sin(\phi) \\
y=y_o +r sin(\theta)sin(\phi) \\
z=z_o + r cos(\phi)
\end{eqnarray}  
with ($0 \le \theta \le 2\pi$ and $0 \le \phi \pi$).
The sensor unit developed takes the geometric information by the object model used and then it computes the needed distances and the unit vectors.

\chapter{Results}

In this chapter we show some case studies selected to explain how the new approach works. The first one shows the improvement obtained with the new task combination function, emphasizing the chattering phenomena absence . In the second one two robots move independently in a room with some moving obstacles. The third one highlights how the Kuka Youbot end effector executes the obstacle avoidance when there is an obstacle placed on its path. The fourth case study, the manipulator end effector is commanded to stay still and a moving spherical obstacle follows a trajectory that would collide with the arm. In the last test, the end effector follows a path to go  to catch a ball in the space: firstly the scene is composed just of the ball afterwards we put in a table and a obstacle, between the ball and the end effector.

\section{First case study: one robot and one obstacle}
As we said in the \emph{Chapter \ref{c:terzo}} we introduce a task combination function to remove the chattering velocity obtained by a \emph{crispy} supervisor. In the case study we want the Kuka Youbot follows a desired path.

\begin{figure}[h]
\centering%
\includegraphics[height=180pt]{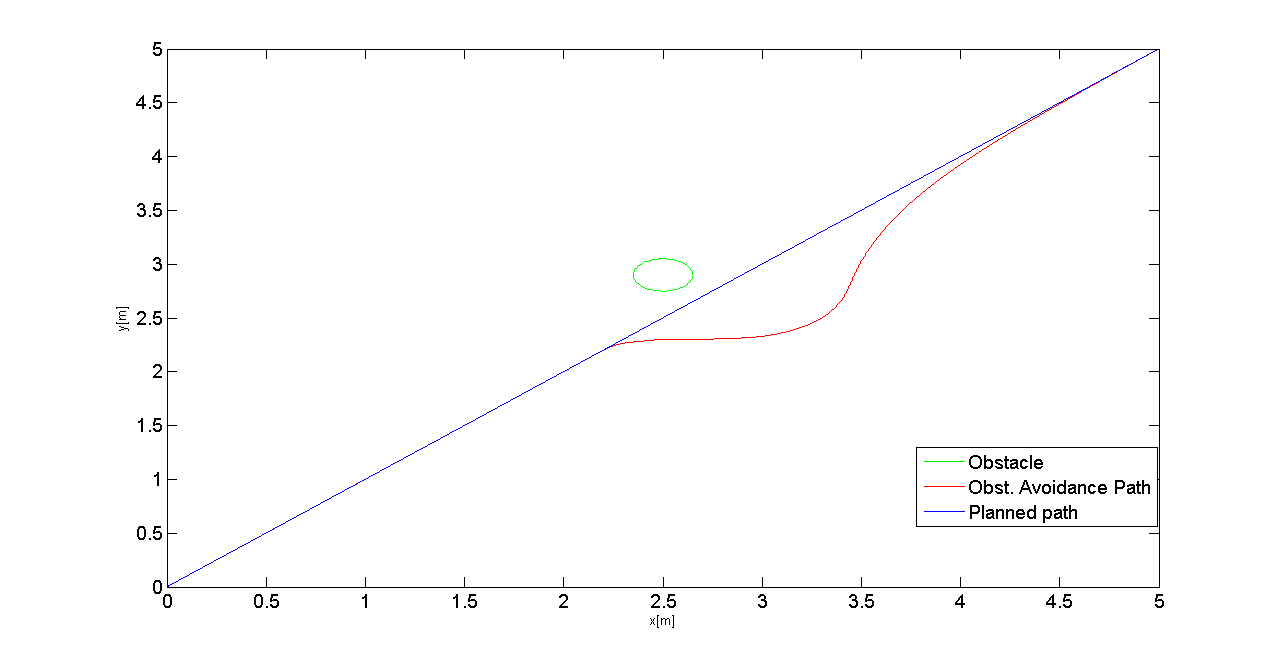}%
\caption{The planned path, and the one obtained with the obstacle avoidance. \label{fig:casouno}}%
\end{figure}

\begin{figure}[h]
\centering%
\includegraphics[height=180pt]{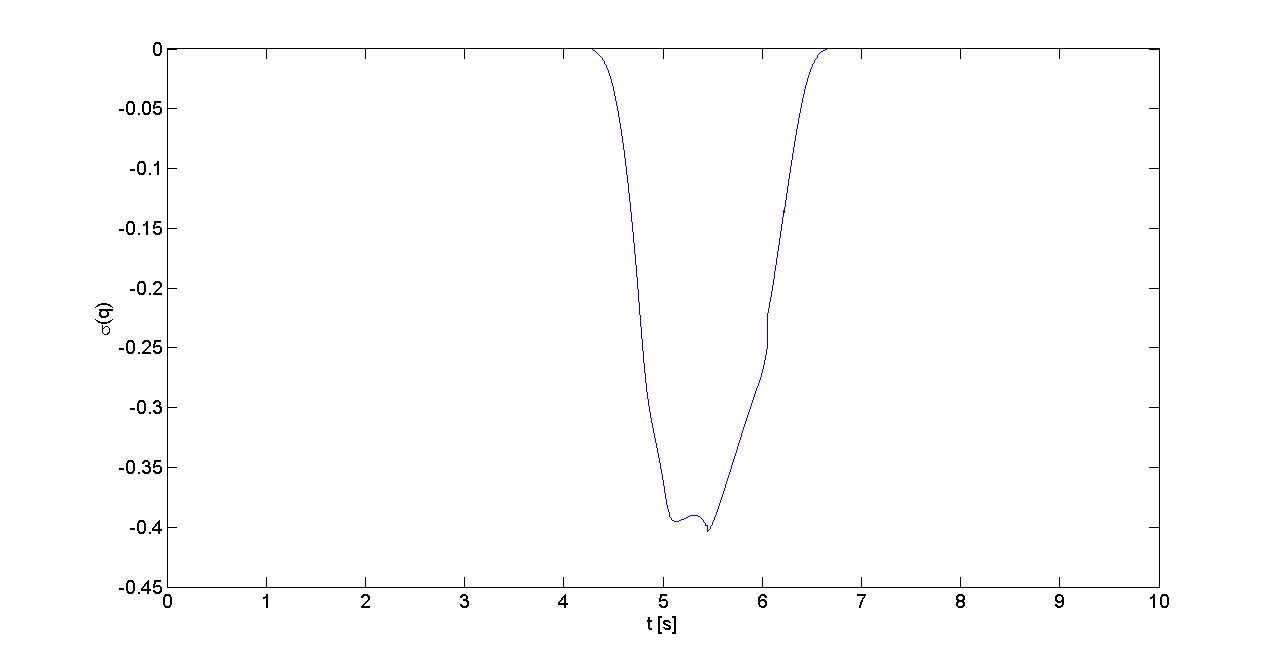}%
\caption{Final history of the pseudo-energy in the first case study. \label{fig:casouno1}}%
\end{figure}

\begin{figure}[h]
\centering%
\includegraphics[height=180pt]{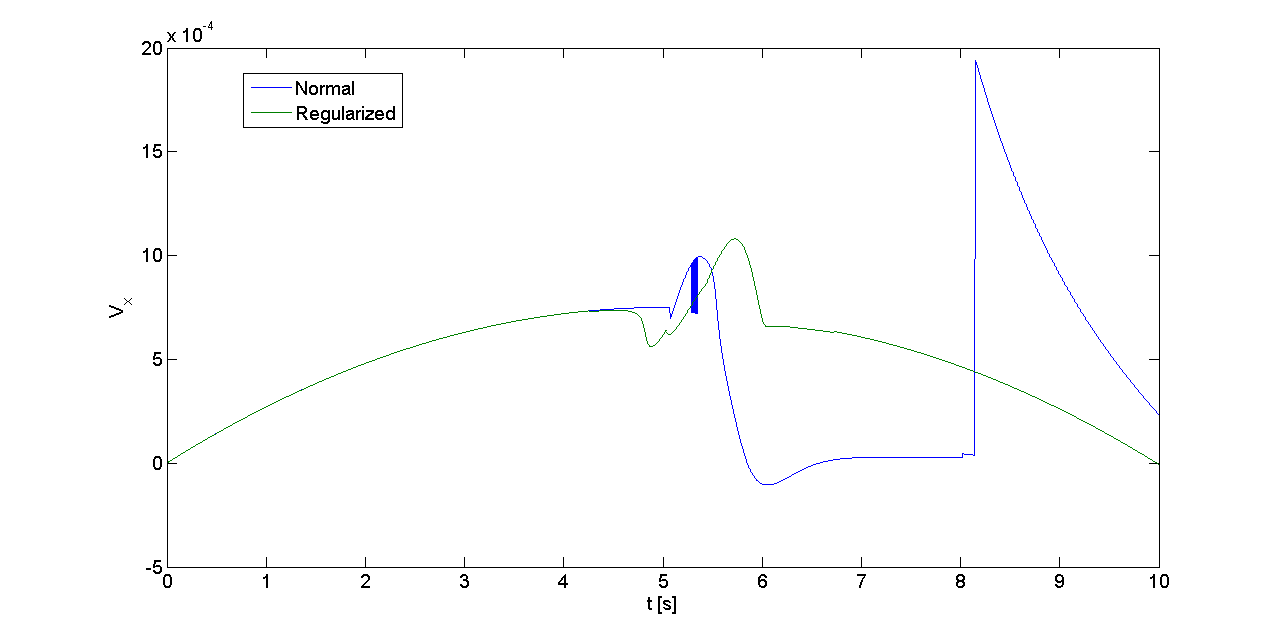}%
\caption{x component of the velocity [$m/s$]. \label{fig:velx}}%
\end{figure}

\begin{figure}[h]
\centering%
\includegraphics[height=180pt]{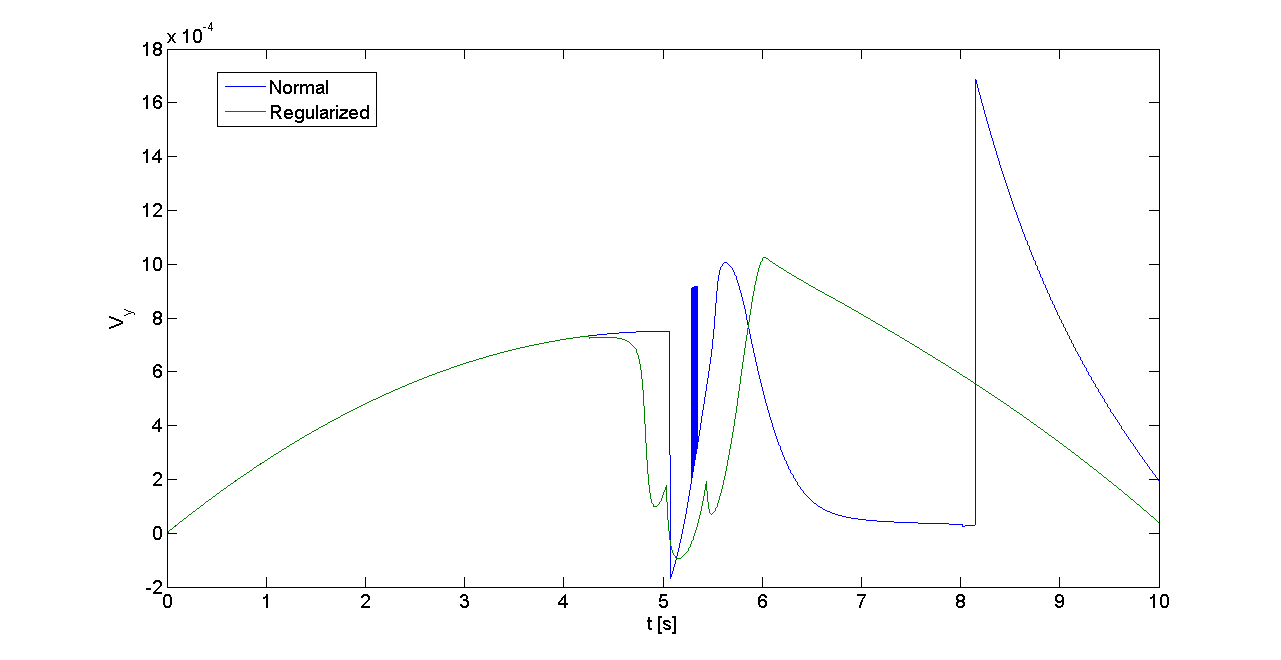}%
\caption{y component of the velocity [$m/s$]. \label{fig:vely}}%
\end{figure}

We placed a spherical obstacle at position $(2.5\hbox{m},2.9\hbox{m})$. The object has a diameter equal to $0.30\hbox{m}$. To pursuit the obstacle avoidance the robot followed the path reported in Figure \ref{fig:casouno}.
For this first case study, we report the parameters values used in the simulations (the sensors are distributed only on the base of the robot):
\begin{itemize}
	\item N=8;
	\item $\gamma_o$=1,$\gamma_g$=2;
	\item $ts=0.001s$;
	\item $f=0.40m$;
	\item $r_k=0.9m$.
\end{itemize}
The pseudo-energy trend associated to the eight sensors is showed in Figure \ref{fig:casouno1}, when the $\sigma$ function is different from zero, the obstacle avoidance behaviour is activated.
Figures \ref{fig:velx} and \ref{fig:vely} report the x and y components of the mobile base velocity obtained with the classical crisp combination law of the two tasks and with the new task combination function. It is evident how the first combination law leads to a chattering velocity, while the proposed method generates smoother velocities. 

\begin{figure}[h]
\centering%
\includegraphics[height=180pt]{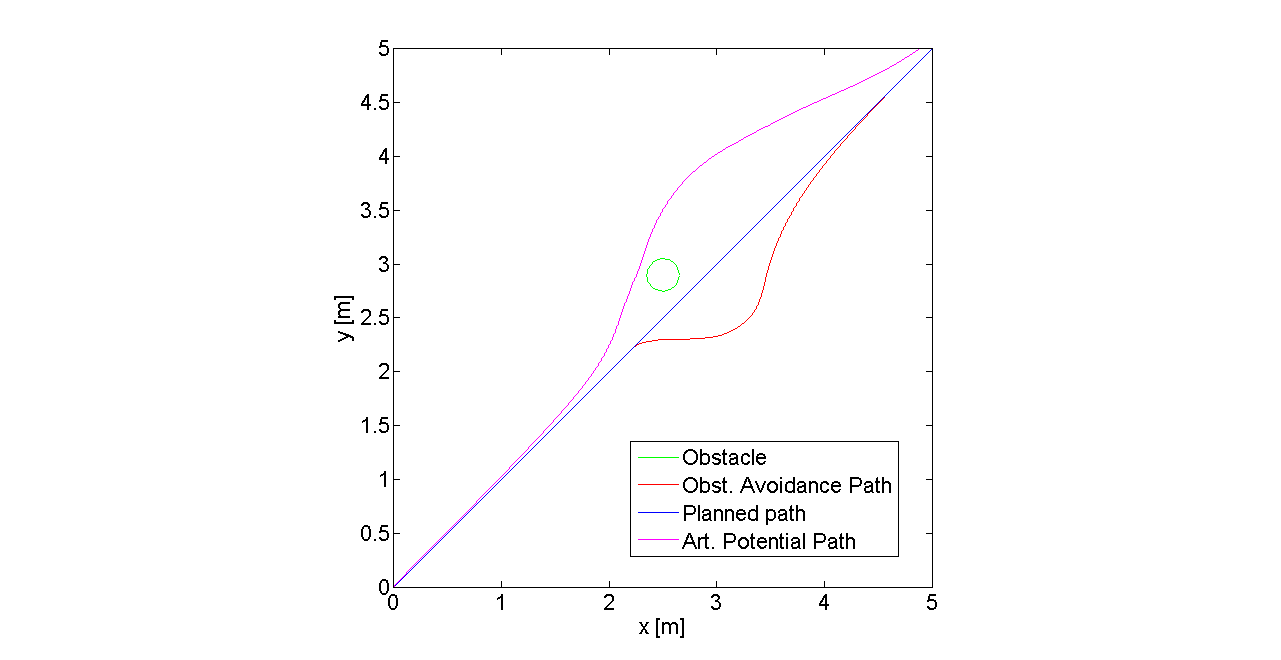}%
\caption{Comparing between our approach and the artificial potential method in the first case study. \label{f:confront}}%
\end{figure}

In this case study we compared the artificial potential method (see Chapter \ref{c:secondo}) with ours, as it is depicted in the Figure \ref{f:confront}, our method responds more quickly then the other one, requiring less time to come back to the original path: because the artificial potential tends to infinity when the robot is near the obstacle, it chooses to follow a path as not close to the  original path as the ours does.

\section{Second case: two robots in a room with moving obstacles}
\begin{figure}[h]
\centering%
\includegraphics[height=180pt]{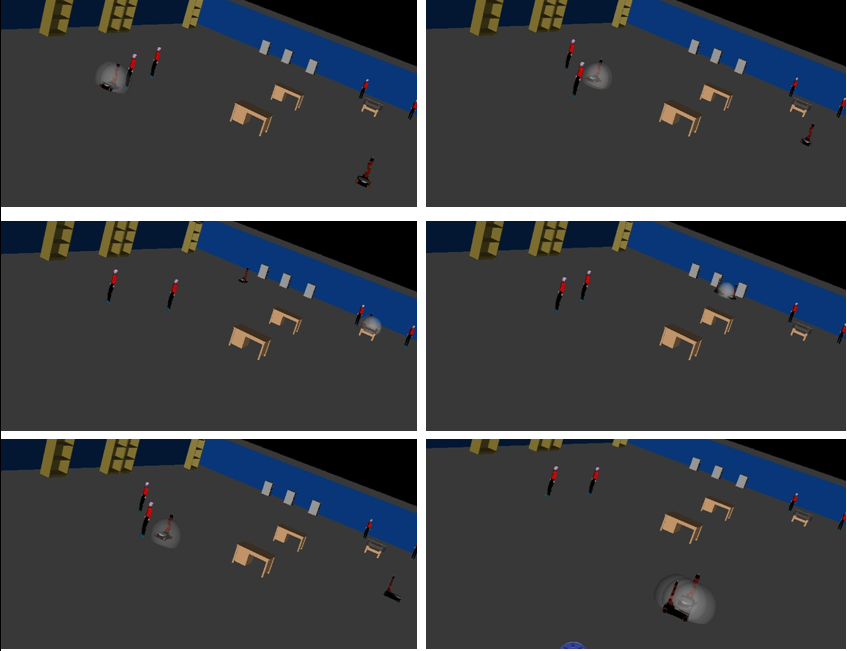}%
\caption{Snapshots of robots moving in a room with multiple obstacle. \label{fig:casodue}}%
\end{figure}

\begin{figure}[h]
\centering%
\includegraphics[height=180pt]{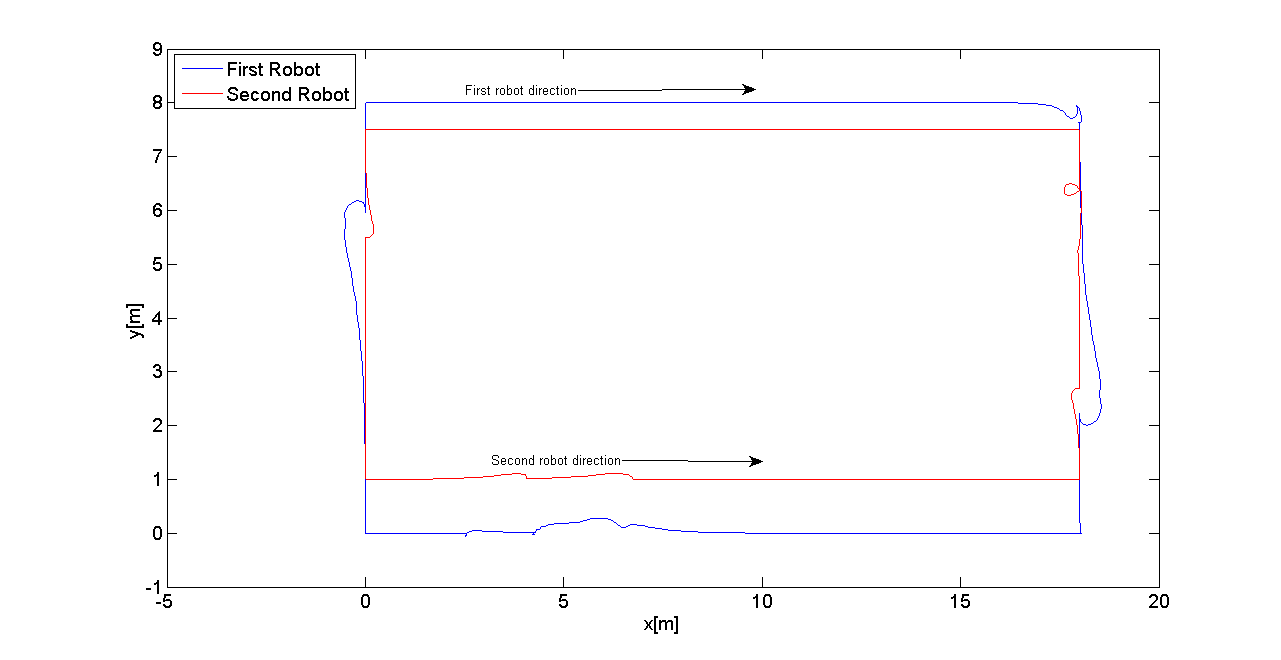}%
\caption{The paths followed by the robots of the second case study. \label{fig:casodue2}}%
\end{figure}

The second case study takes into account two Kuka Youbot robots moving separately on two different paths. For each robot there is a control unit, both sharing the same environment. The Figure \ref{fig:casodue} is a collection of  this case's snapshots. The time sequence is from left to right and from top to bottom.
In this case study the parameters values are the same as reported in the first one, for both robots.
The Figure \ref{fig:casodue2} shows the obtained paths, which show that significant adjustments are needed to avoid collision with humans, fixed obstacles and other robots.
In the environment we put not only fixed obstacles (desks and bookshelves) but also some moving obstacles  (human beings, to simulate a possible real case where mobile robots can be used for surveillance). 

\section{Third case study: Obstacle on the end-effector path}
\begin{figure}[h]
\centering%
\includegraphics[height=180pt]{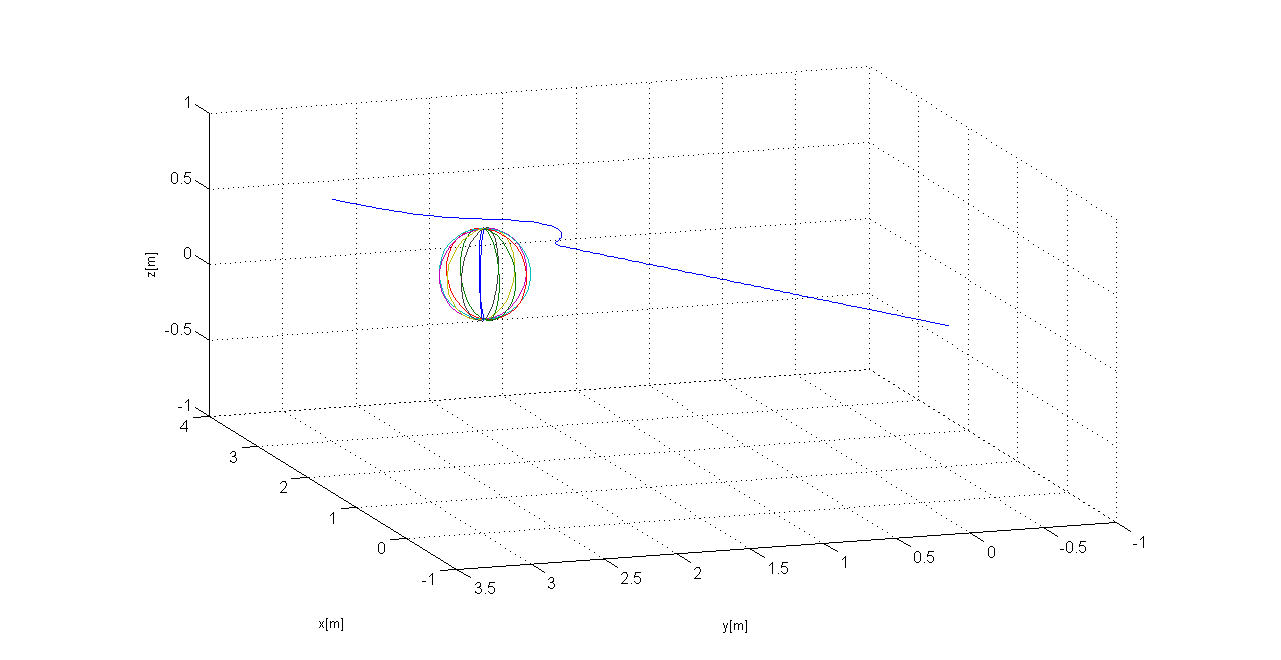}%
\caption{The modified path and the obstacle of the third case study. \label{fig:ost}}%
\end{figure}

\begin{figure}[h]
\centering%
\includegraphics[height=180pt]{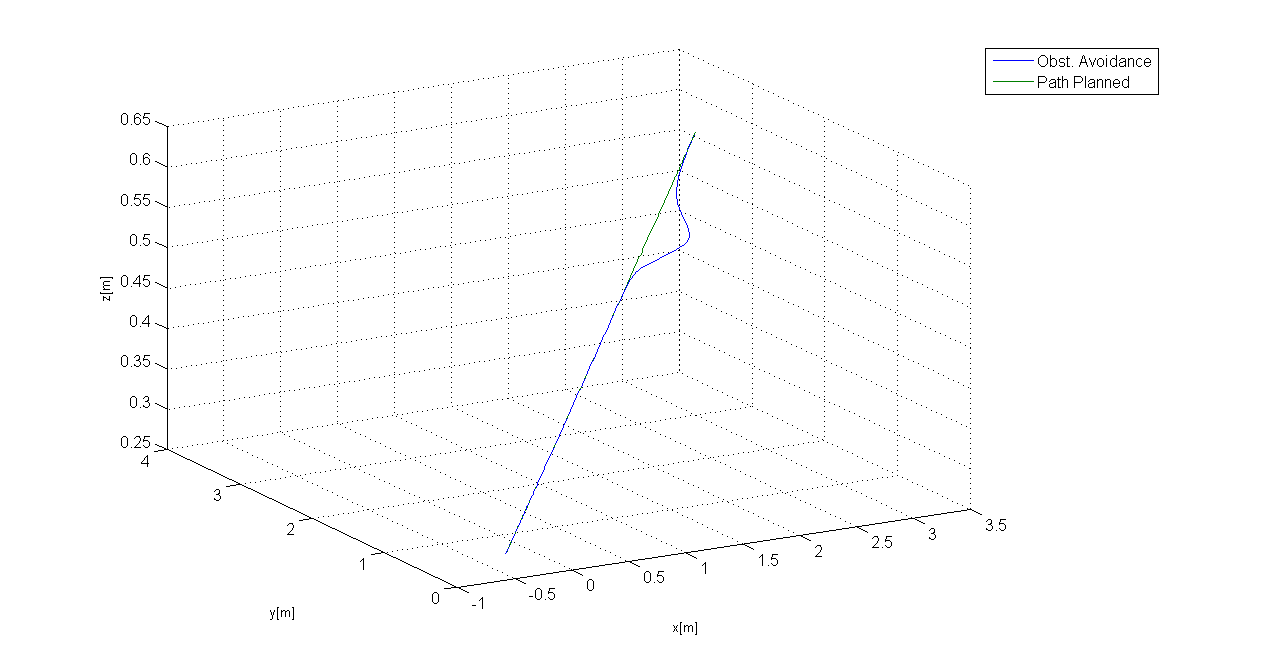}%
\caption{The modified and planned paths of the third case study. \label{fig:ost1}}%
\end{figure}

In this case study, an obstacle is situated on a line segment path followed by the Kuka Youbot end effector. The sensors are placed also on the manipulator links, thus the $f$ are smaller then the previous ones, $\gamma_o$ and $\gamma_g$ have the same values as the previous case studies.  The Figure \ref{fig:ost} shows the modified path and the obstacle. In Figure \ref{fig:ost1} we plotted the modified path and the planned one. 
Both figures clearly show that the end-effector path is locally adjusted to avoid the obstacle and to verify that no collision occurs with any part of the robot.

\begin{figure}[h]
\centering%
\includegraphics[height=180pt]{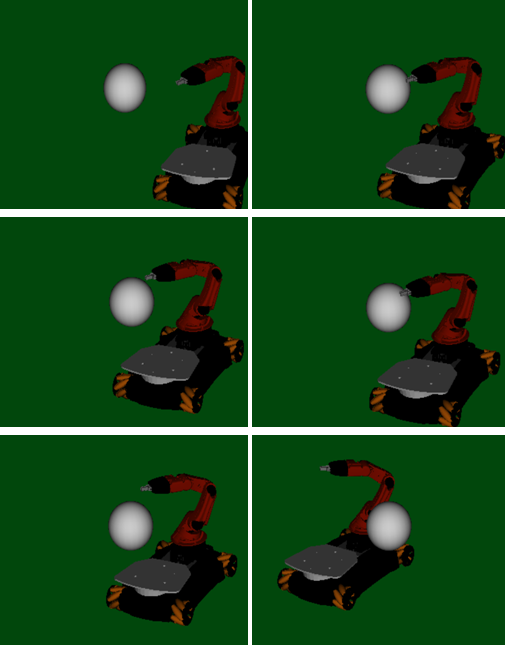}%
\caption{Snapshots of the robot motion in the third case study. \label{fig:ost2}}%
\end{figure}

Figure the \ref{fig:ost2} is a snapshots collection of the manipulator movements. 

\section{Fourth case study: A still manipulator and a moving obstacle}

\begin{figure}[h]
\centering%
\includegraphics[height=180pt]{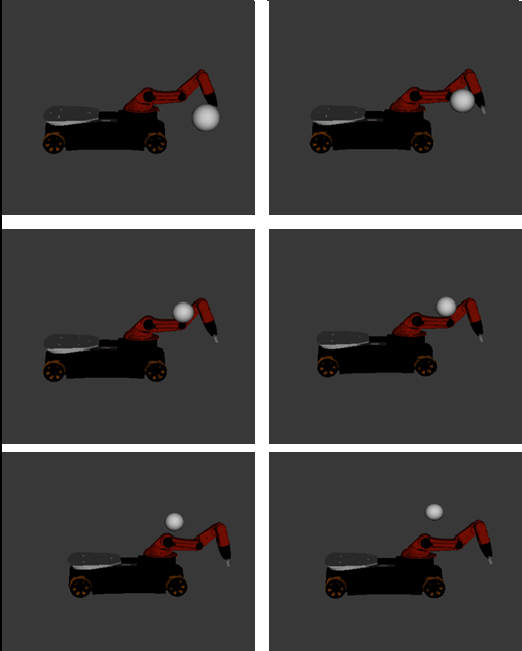}%
\caption{Snapshots of the robot motion in the fourth case study. \label{fig:ost21}}%
\end{figure}

\begin{figure}[h]
\centering%
\includegraphics[height=125pt]{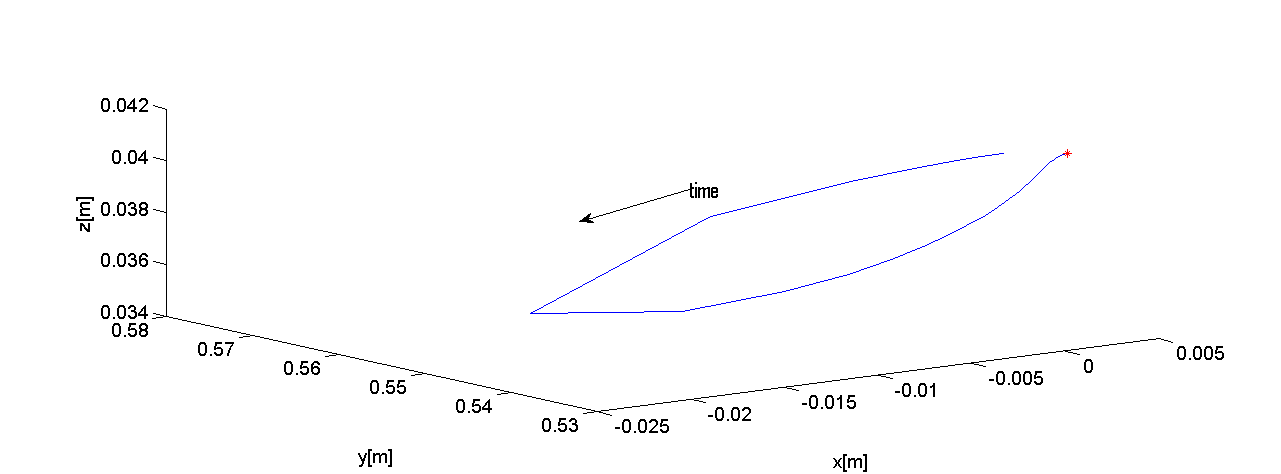}%
\caption{The end effector path of the fourth case study. \label{fig:ost22}}%
\end{figure}

\begin{figure}[!h]
\centering%
\includegraphics[height=180pt]{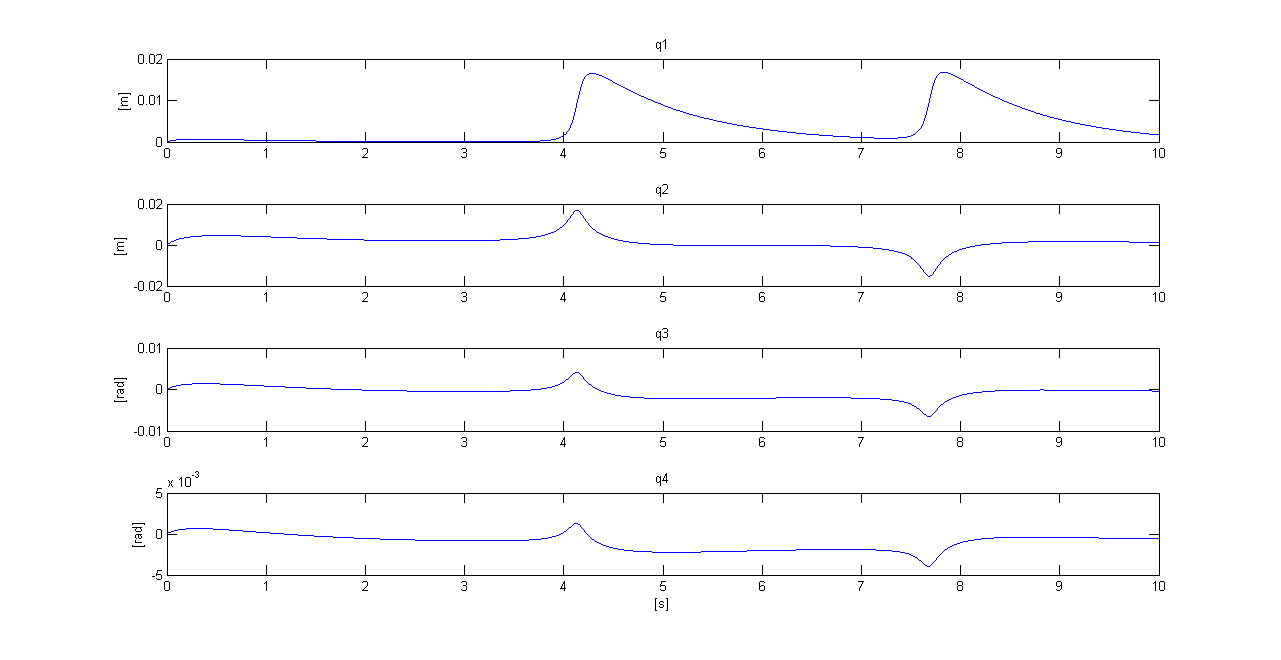}%
\caption{The first four configuration variables of the fourth case study. \label{fig:primi}}%
\end{figure}

\begin{figure}[!h]
\centering%
\includegraphics[height=180pt]{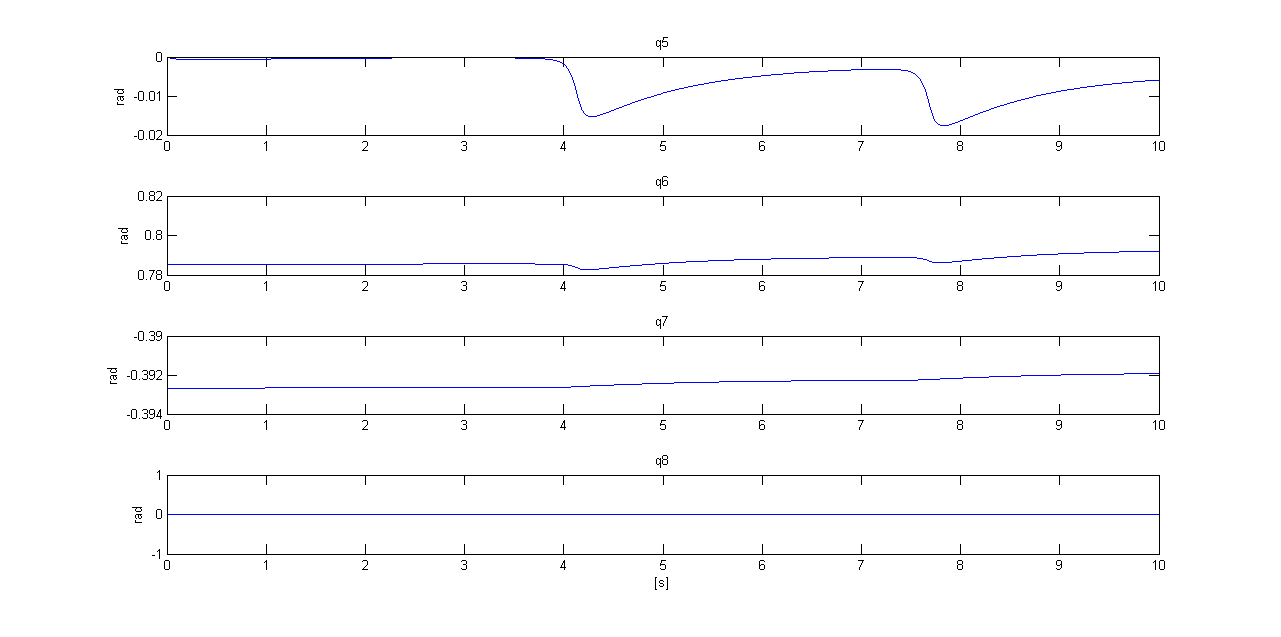}%
\caption{The second four configuration variables of the fourth case study. \label{fig:secondi}}%
\end{figure}

In this case study a moving obstacle tries to hit a robot: the Kuka Youbot avoids the obstacle and try to keep the end effector in the desired position. Sixteen sensors are mounted on the base and on the manipulator of the robot, the $f$s relative to the manipulator's sensor points are smaller then the ones associated to the sensors on the base. Also for this case the values of $\gamma_o$ and  $\gamma_g$ are equal to 1. The snapshots reported in Figure \ref{fig:ost21} show that the robot moves so as to avoid the obstacle trying to keep the end effector at the desired position. Note that in the case of multiple non-conflicting tasks the NSB does not guarantee that the lower priority task is instantaneously achieved with the sub-optimal velocity. Nevertheless, in the considered case, the closed loop ensures that the error of secondary task converges to zero. This means that the end-effector returns to the desired position only after a transient of the motion needed to avoid the obstacle. This is proved by the path actually followed by the end effector reported in Figure \ref{fig:ost22} which returns to the desired location indicated by the red star. Furthermore we report the configuration variables $q=[q1,q2,\cdots,q8]$ obtained by the executed simulation, they are displayed in Figures \ref{fig:primi} and \ref{fig:secondi}.

\section{Fifth case study: picking a ball on a table}

\begin{figure}[h]
\centering%
\includegraphics[height=180pt]{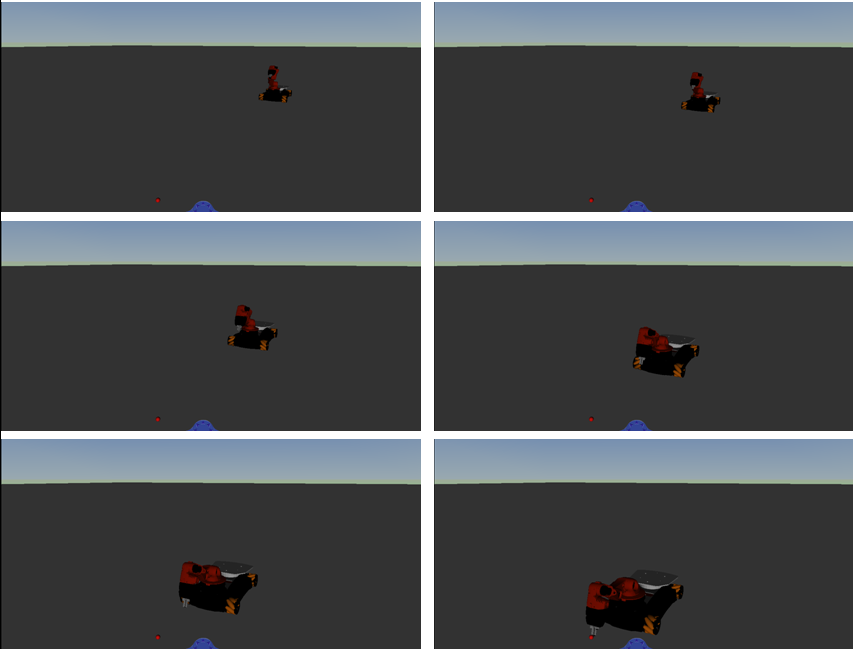}%
\caption{Snapshots of the robot motion in the fifth case study, without considering obstacles. \label{fig:ost31}}%
\end{figure}

\begin{figure}[h]
\centering%
\includegraphics[height=180pt]{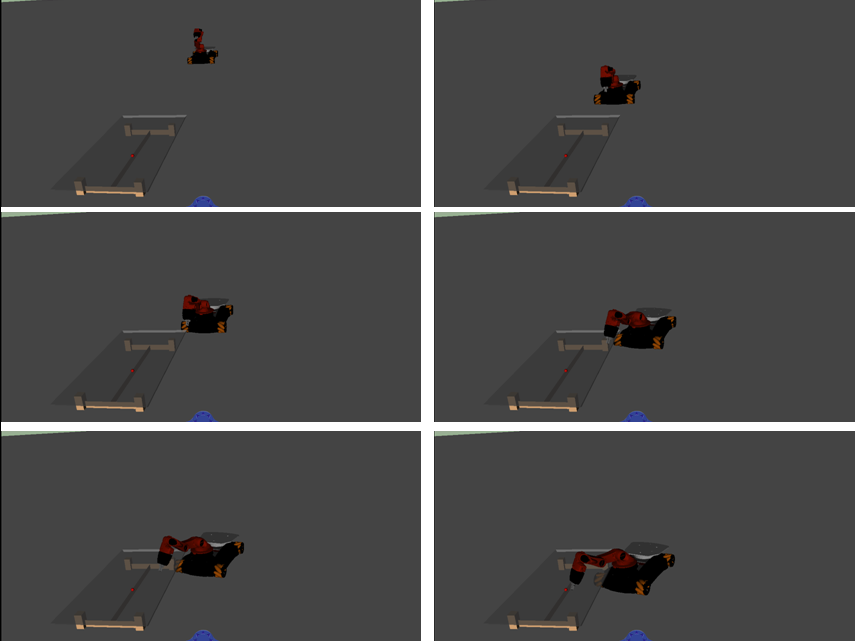}%
\caption{Snapshots of the robot motion in the fifth case study, with object on the table. \label{fig:ost32}}%
\end{figure}

\begin{figure}[h]
\centering%
\includegraphics[height=180pt]{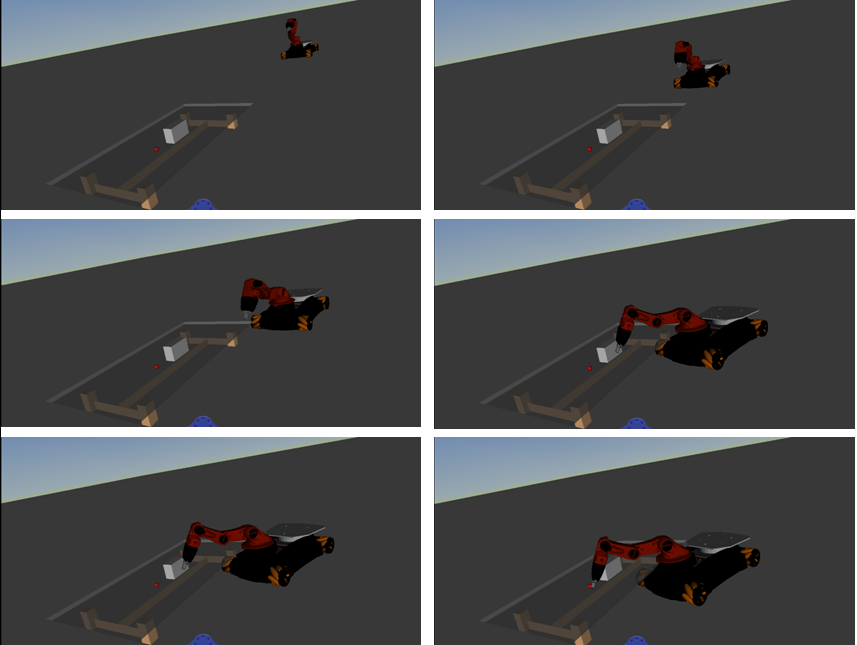}%
\caption{Snapshots of the robot motion in the fifth case study, with objects on the table and a small box. \label{fig:ost33}}%
\end{figure}

In this test a robot has to follow the planned path. From the point $pi=(0\hbox{m},0.51\hbox{m},0.30\hbox{m})$ to the point $pf=(-0.03\hbox{m},3.2\hbox{m},0.02\hbox{m})$ passing through the point $pf_1=(-0.03\hbox{m},3\hbox{m},0.05\hbox{m})$ using two line segments. In the point $pf$ we put a ball to be picked up.  In the scene we put, gradually, two obstacles: a table and a small box. The table is centred in the scene in a way to have the ball on it. The small box is put between the point $pf$, and $pf_1$. For this case study we implemented the task combination function written in (\ref{e:linear}).
Figures \ref{fig:ost31}, \ref{fig:ost32}, \ref{fig:ost33} are a collection of snapshots for the case in study.

\begin{figure}[h]
\centering%
\includegraphics[height=180pt]{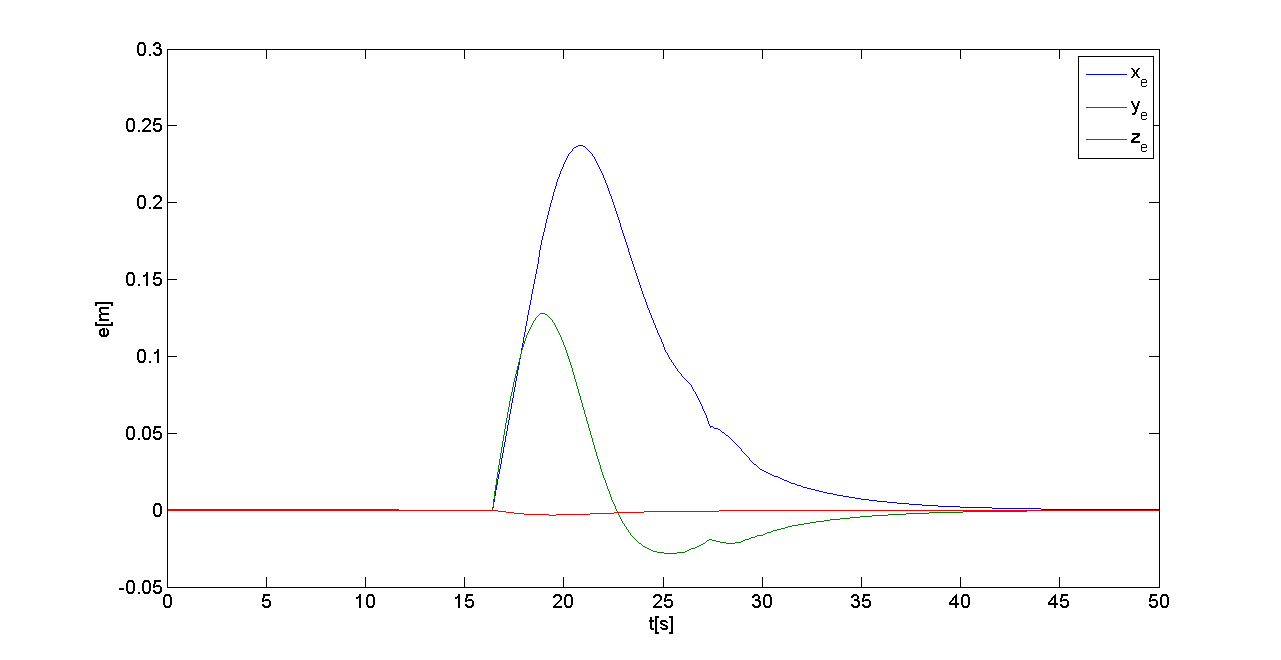}%
\caption{The end effector position error in the fifth case study (with the table and the box).  \label{fig:error}}%
\end{figure}

\begin{figure}[h]
\centering%
\includegraphics[height=180pt]{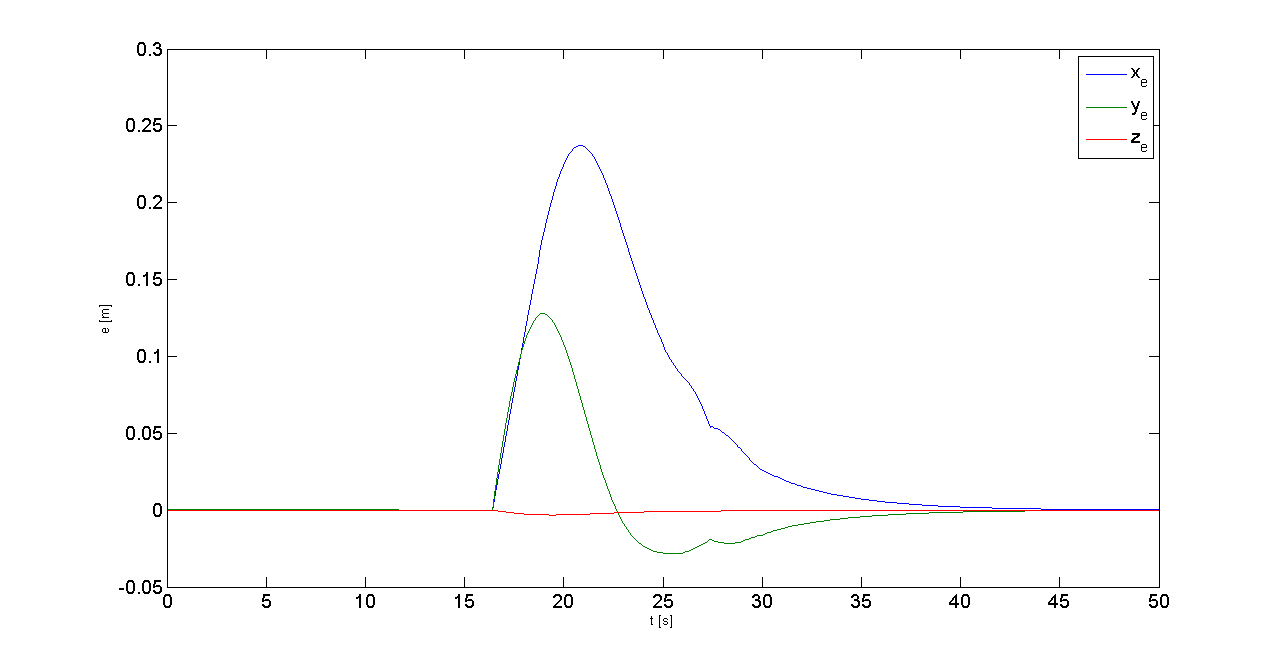}%
\caption{The end effector position error in the fifth case study (with only the table)  \label{fig:error1}}%
\end{figure}

\begin{figure}[h]
\centering%
\includegraphics[height=180pt]{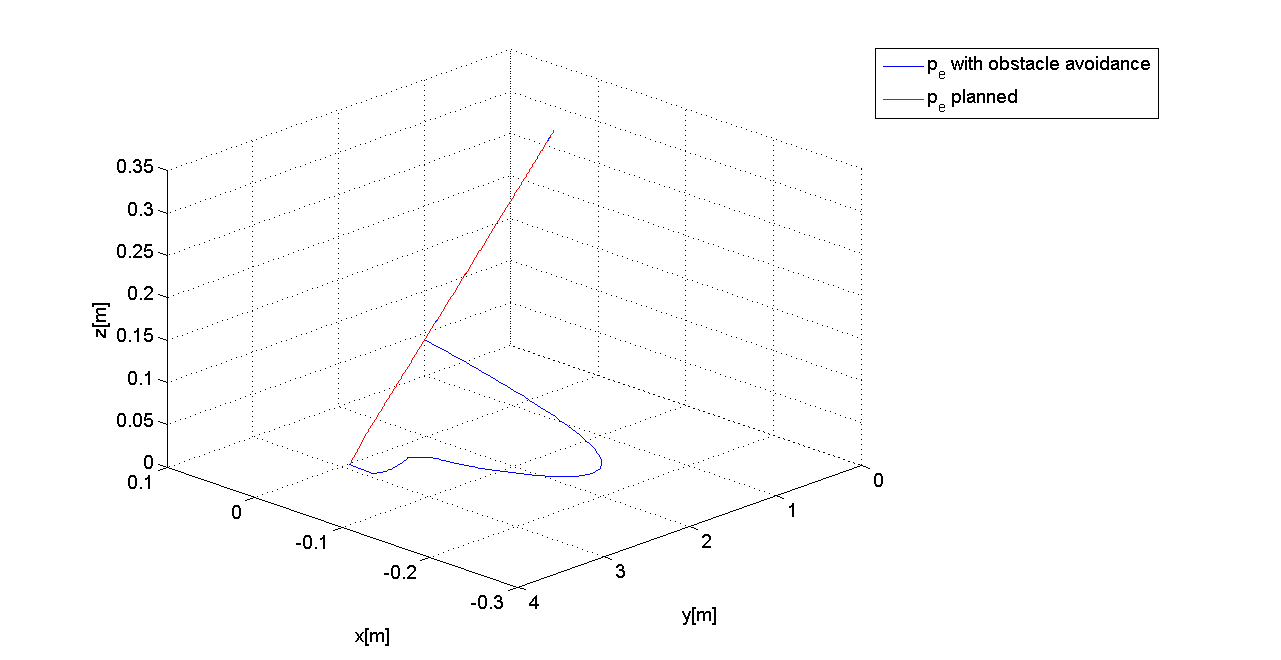}%
\caption{The paths in the fifth case study (with only the table).  \label{fig:pathx1}}%
\end{figure}

\begin{figure}[h]
\centering%
\includegraphics[height=180pt]{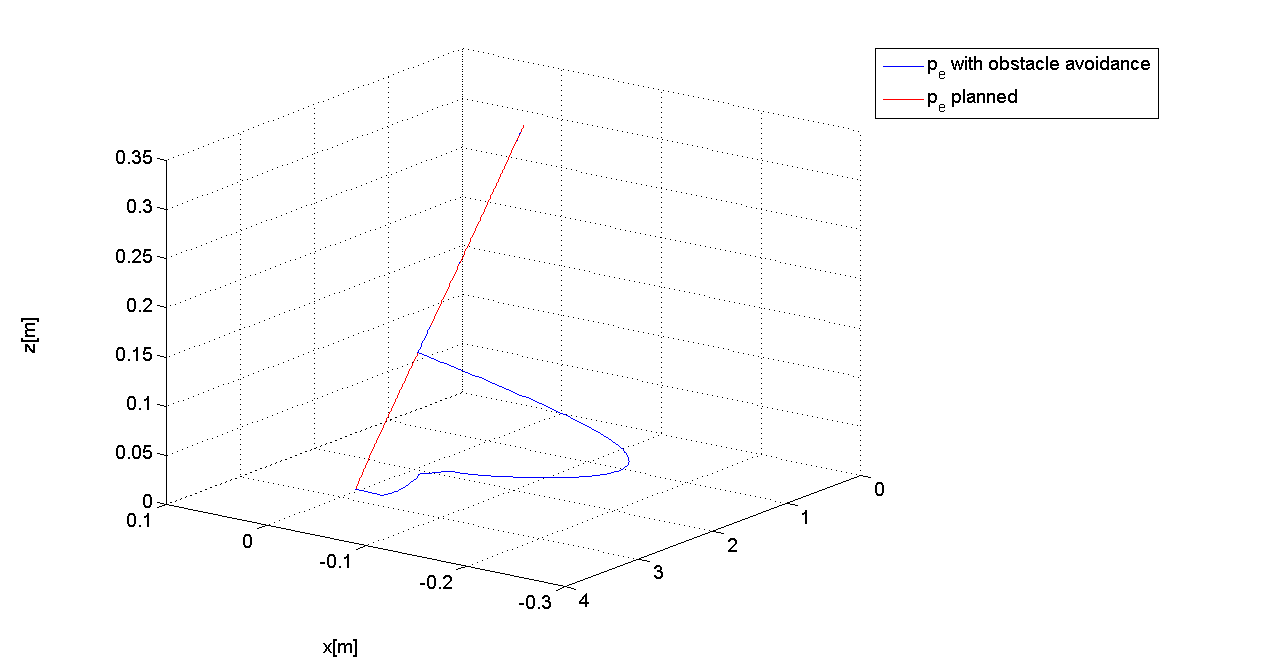}%
\caption{The paths in the fifth case study (with table and wall).  \label{fig:pathx2}}%
\end{figure}

Figure \ref{fig:error} shows the CLIK convergence in terms of the end-effector position error with the table and the small box. Figure \ref{fig:error1} reports the error obtained in a simulation where there is only the table and the ball. In both cases, we plot the end effector path: the planned one and the one computed by the obstacle avoidance algorithm (Figures \ref{fig:pathx1} and \ref{fig:pathx2}) .

In this case study we used the following parameters:
\begin{itemize}
	\item \textit{N}=17, we added a sensor on the top of the end effector, we turn it off when the distance between the end effector and the ball is less then $5\hbox{cm}$;
	\item $r_k$=5m, for each sensor;
	\item the $f$ are: 0.35\textit{m} for the sensors on the base, 0.3\textit{m} for the sensors on the first link, 0.1\textit{m} the ones on the second link and 0.02\textit{m} the one mounted on the end effector;
	\item $\gamma_g$ is 2;
	\item $\gamma_o$ is 1;
	\item $t_s$ is 0.01\textit{s};  
	\item task combination parameter \textit{$\epsilon$} is 0.08\textit{m} (see equation (\ref{e:linear})).
\end{itemize}
The use of a linear task combination function lets the robot move in a more fluid way then the sigmoidal one. The chattering phenomena can occur if the \textit{$\epsilon$} parameter is too small.

\newpage

\chapter{Conclusions and future developments}

The new approach satisfy to the requirements listed in \emph{Chapter \ref{c:terzo}}. The method proposed in this work to address the obstacle avoidance problem for a mobile manipulator exploits information provided by distributed proximity perception so as to avoid collision of all the parts of the robot with objects as well as self-collisions. The algorithm requires very limited a priori knowledge about the 3D model of the robot and no information on the environment and thus it is suitable for robotic tasks executed in a dynamic, unstructured environment. There is no a standard analytics procedure to place the sensors on the robots, you can place them randomly. We suggest to cover all the part of the robot to protect. 

For its non-conservative approach it can be used on humanoid robots, to permit them to interact with the dynamically world in a safer and autonomous way.

This objective has been achieved by adopting a behaviour based control approach with two key innovations. 

First, the definition of the pseudo-energy only on the basis of the distributed proximity perception and it does not require visual information. This strategy can be physically interpreted as a sort of elastic buffer covering the parts of the robot around each sensor point when this point comes close to an obstacle.

The second innovation is the introduction of the concept of the task combination function, that consists in a more flexible and general way to combine multiple tasks in order to obtain a smooth motion of the robot avoiding chattering
phenomena. 

Another feature of the method is the strict coordination between the base and the arm exploiting the redundant degrees of freedom, a relevant topic in mobile manipulation.

The differences with other methods can be summarized as follows:
\begin{itemize}
    \item the distributed management of the control points is similar to that of the artificial potential  but, since we define the priority of the tasks there is no need to define the potential that tend to infinity when the robot is near the obstacle, it means we need less computation also because the information of all the sensors is not always needed, we need only the information of the sensors which are activated compressed springs ;
\item we guarantee an autonomous handling of task priorities thanks to the presence of the  NSB approach, and differently from the classical NSB we defined a regularized supervisor, the task combination function to avoid a chattering robot velocity.
\end{itemize}

Obviously we expect to study some more use cases to prove the potentiality of the described approach, especially for humanoid robots.

As possible future developments, a first extension is to consider the case of multiple robots. Owing to the NSB approach, the coordination among robots of a team can be achieved by introducing an additional behaviour.

Another interesting future line of research is the extension of the behavioural approach based on distributed perception to the physical interaction between robots and the environment, possibly including humans. This would require more sophisticated sensing systems able to provide tactile information but the method to handle such data could be the same proposed here.

\end{document}